\documentclass[11pt]{article}
 
\usepackage[a4paper,top=2.6cm,bottom=2.6cm,left=2.5cm,right=2.5cm]{geometry}
\usepackage{setspace}
\usepackage[utf8]{inputenc}
\usepackage[T1]{fontenc}
\usepackage{lmodern}
\usepackage[english]{babel}
\usepackage{microtype}
 
\usepackage{amsmath}
\usepackage{amssymb}
 
\usepackage{graphicx}
\usepackage{booktabs}
\usepackage{array}
\usepackage{float}
\usepackage{caption}
\usepackage{colortbl}
\usepackage[dvipsnames]{xcolor}
\usepackage[numbers,sort&compress]{natbib}
\usepackage[colorlinks=true,linkcolor=NavyBlue,citecolor=NavyBlue,
            urlcolor=NavyBlue]{hyperref}
\usepackage{enumitem}
\setlist{itemsep=2pt,topsep=4pt}
 
\usepackage{tikz}
\usetikzlibrary{arrows.meta,positioning,fit,backgrounds,calc,shapes.geometric}
\definecolor{figline}{gray}{0.35}
\definecolor{figfill}{gray}{0.96}
\definecolor{figaccent}{gray}{0.88}
\tikzset{
  fnode/.style   = {draw=figline, line width=0.4pt, rounded corners=2pt,
                    fill=white, align=center, inner sep=4pt, font=\footnotesize},
  snode/.style   = {fnode, fill=figfill},
  anode/.style   = {fnode, fill=figaccent},
  gnode/.style   = {draw=figline, line width=0.4pt, dashed,
                    rounded corners=3pt, inner sep=7pt},
  farrow/.style  = {-{Latex[length=4pt,width=3pt]}, draw=figline, line width=0.4pt},
  floop/.style   = {farrow, dashed},
  flab/.style    = {font=\scriptsize\itshape, text=figline},
  fsub/.style    = {font=\scriptsize, text=figline},
}
 
\definecolor{oursbg}{gray}{0.91}
\newcommand{\ours}{\rowcolor{oursbg}}

\newcommand{\clmmlm}{CLM$\rightarrow$MLM}
 
\usepackage{titlesec}
\titleformat{\section}{\normalfont\large\bfseries}{\thesection}{0.6em}{}
\titleformat{\subsection}{\normalfont\normalsize\bfseries}{\thesubsection}{0.6em}{}
\titlespacing*{\section}{0pt}{1.4em}{0.6em}
\titlespacing*{\subsection}{0pt}{1.0em}{0.4em}
 
\title{\textbf{MoganBert-TR: A Turkish Encoder Foundation Model\\
Trained from Scratch with a \clmmlm{} Curriculum}}
 
\author{
Furkan Yılmaz \\ \texttt{furkanyl509@gmail.com}
\and
Habibe Aleyna Taşdemir \\ \texttt{aleynattasdemir@gmail.com}
\and
Muhammed Faruk Gözay \\ \texttt{gozayfaruk@gmail.com}
}
\date{}
 
\begin{document}
\maketitle
 
\begin{abstract}
\noindent
Publicly available large-scale pretraining data for Turkish exists largely as a
subset of multilingual dumps and involves no language-specific quality
filtering; existing Turkish encoders, meanwhile, have adopted modern
architectures while leaving the pretraining objective fixed. This paper
introduces an encoder foundation model trained from scratch on a
language-specifically filtered corpus (MoganBert-TR, 149M parameters), together
with an embedding model derived from it (MoganBert-Embed).

\textbf{(1) Training objective.} MoganBert-TR is trained over 237.3 billion
tokens with a two-stage \clmmlm{} curriculum: the first portion of the run uses
causal language modelling and the remainder masked language modelling, with the
transition made inside the stable phase of a WSD schedule and without touching
the learning rate. In a controlled ablation on the same architecture and the
same corpus under an equal step budget, this design outperforms pure MLM by
2.7--3.7$\times$ on Turkish MS MARCO retrieval; the measured mechanism is
embedding geometry a single direction absorbs 28.1\% of the variance in the
pure-MLM model, against 11.9\% under the curriculum.
\textbf{(2) Branched annealing.} Long-context extension and learning-rate decay
are split into two branches after a shared prefix; running the final portion of
decay at 1024 context improves the TrGLUE average by $+0.49 \pm 0.26$ points
across five paired seeds ($p \approx 0.013$) and beats a model-soup alternative
by 0.75 points at $\sim$4.3\% additional cost.
\textbf{(3) Results.} MoganBert-TR attains 78.41 on TrGLUE, the best among the
Turkish ModernBERT models compared here, and 77.73 on TabiBench, second among
monolingual Turkish encoders and 0.19 points behind the leader; it leads two of
the eight TabiBench categories, with the largest margin on code retrieval
($+3.62$ points over TabiBERT). \textbf{(4) Embedding model.}
MoganBert-Embed, produced through teacher distillation and multi-signal
contrastive fine-tuning, ranks first among student models on the MTEB(Turkish)
overall average with 68.30 and reaches 99.5\% of its 7.57-billion-parameter
teacher's score with a 51$\times$ smaller backbone. \textbf{(5) Data and
tokenizer.} The Turkish portion of FineWeb2, recent months pulled from raw
Common Crawl, and domain-dense text extracted from printed and institutional
sources with a vision-language model were combined into a single pipeline;
since no off-the-shelf Turkish quality classifier exists, the decisions of a
fine-tuned Turkish BERT were distilled into fastText (94.4\% agreement,
$\sim$90$\times$ speed-up). The accompanying 50,048-token tokenizer outperforms
all compared Turkish tokenizers on both compression and fertility across two
independent test sets.

Model weights, tokenizer, embedding model and evaluation code will be released
openly at \url{https://huggingface.co/moganai}.
\end{abstract}

\vspace{0.5em}

\section{Introduction}
\label{sec:intro}

Despite the visibility of generative models, encoder-only models remain the de
facto backbone of classification, sequence labelling and retrieval pipelines:
their inference cost is low, they see bidirectional context in a single pass,
and they are natural candidates for producing embeddings.
ModernBERT~\citep{modernbert} updated the architectural side of this family with
RoPE~\citep{rope}, alternating local/global attention and GLU; the
language-specific adaptations that followed demonstrated that the architecture
transfers beyond English.

On the Turkish side this transfer has been carried out only partially.
BERTurk~\citep{berturk} long served as the de facto standard, and models such as
TabiBERT~\citep{tabibert} and ModernBERT-TR~\citep{modernberttr} have brought
the ModernBERT architecture to Turkish. What these works share, however, is the
assumption that the pretraining objective is fixed: pure MLM. While the
architecture was being modernised, the \emph{training objective} itself was
never systematically tested for Turkish.

This gap is not trivial. Controlled comparisons of encoder pretraining
objectives~\citep{clmmlm} report that, under a fixed compute budget, a
two-stage \clmmlm{} design outperforms pure MLM; yet that finding was produced
at English scale, and whether it holds for a morphologically rich, agglutinative
language with its own tokenizer and its own corpus remains an open
question. Likewise, \emph{how} long-context extension should be carried out
during the annealing phase has no established recipe: mmBERT~\citep{mmbert}
shows that altering the data mixture of the decay phase is decisive, but the
role of context length within that same phase has not been measured.

This paper addresses both questions through MoganBert-TR, an encoder foundation
model trained from scratch for Turkish (149M parameters). The model is trained
over 237.3 billion tokens with a two-stage \clmmlm{} curriculum; the
contribution of that design is measured through a controlled ablation on the
same architecture and the same corpus; and the annealing phase is branched in
order to test the interaction between long-context extension and learning-rate
decay across paired seeds. Training uses a language-specifically filtered
corpus; the data pipeline and the tokenizer were produced for this purpose and
are transferable contributions in their own right.

\paragraph{Contributions.}
\begin{itemize}[leftmargin=1.4em]
\item \textbf{A training-objective ablation for Turkish.} Under an equal step
budget, \clmmlm{} is tested against pure MLM. While the two designs are
practically tied on linear probing tasks, the curriculum leads by
2.7--3.7$\times$ on Turkish MS MARCO retrieval. The mechanism is attributed to
embedding geometry: a single direction absorbs 28.1\% of the variance in the
pure-MLM model, against 11.9\% under the curriculum
(Section~\ref{sec:ablation}).

\item \textbf{Branched annealing.} Long-context extension and learning-rate
decay are split into two branches after a shared prefix. Running the final
portion of decay at 1024 context improves the TrGLUE average by
$+0.49 \pm 0.26$ points across five paired seeds ($p \approx 0.013$; all five
seeds positive) and beats a model-soup alternative by 0.75 points at
$\sim$4.3\% additional cost (Section~\ref{sec:branches}).

\item \textbf{MoganBert-TR.} With 78.41 on TrGLUE it is the best of the Turkish
ModernBERT models compared here; on TabiBench it scores 77.73, second among
monolingual Turkish encoders behind ModernBERT-TR (77.92) and ahead of
TabiBERT (77.58), and leads two of the eight categories
(Section~\ref{sec:tabibench}). The 1.46-point margin by which BERTurk
leads is concentrated almost entirely in the known weak spots of an NSP-free
architecture (CoLA, STS-B), analysed separately
(Section~\ref{sec:cola}).

\item \textbf{MoganBert-Embed.} The embedding model produced through teacher
distillation and multi-signal contrastive fine-tuning ranks first among student
models on the MTEB(Turkish) overall average with 68.30 and reaches 99.5\% of
its 7.57-billion-parameter teacher's score with a 51$\times$ smaller backbone
(Section~\ref{sec:embed}).

\item \textbf{Data pipeline and tokenizer.} A distillation chain that, in a
language without an off-the-shelf quality classifier, uses a slow but accurate
teacher as a label generator (94.4\% agreement, $\sim$90$\times$ speed-up), and
a 50,048-token tokenizer that outperforms all compared Turkish tokenizers
across two independent test sets (Sections~\ref{sec:data}
and~\ref{sec:tokenizer}).

\item \textbf{Open release.} Model weights, tokenizer, embedding model and
evaluation code will be released at \url{https://huggingface.co/moganai}.
\end{itemize}

The remainder of the paper is organised as follows: Section~\ref{sec:related}
reviews prior work; Section~\ref{sec:data} introduces the pretraining data and
Section~\ref{sec:tokenizer} the tokenizer; Section~\ref{sec:model} presents the
model architecture, the training-objective ablation and the full pretraining
run; Section~\ref{sec:eval} reports evaluation results and
Section~\ref{sec:embed} the embedding model; Section~\ref{sec:discussion}
discusses the findings, Section~\ref{sec:limitations} lists limitations, and
Section~\ref{sec:conclusion} concludes. Implementation details are given in the
appendices.

\section{Related Work}
\label{sec:related}

\subsection{The evolution of encoder architectures}
The encoder-only paradigm established by BERT~\citep{bert} has been updated over
subsequent years by a series of efficiency- and stability-oriented
improvements. ModernBERT~\citep{modernbert} consolidates this accumulation into
a single recipe: long-context support via RoPE~\citep{rope}, alternating
local/global attention, GLU activations, unpadding and FlashAttention-based
kernels. The result is an 8192-token context and a marked inference speed-up.
This architecture has also become the de facto basis for language-specific
adaptations.

\subsection{Turkish encoder models}
BERTurk~\citep{berturk}, the first monolingual BERT trained from scratch for
Turkish, long served as the de facto standard; because it is not the product of
an academic publication, its corpus composition and preprocessing decisions
remain partly unspecified. TabiBERT~\citep{tabibert} is the first work to bring
the ModernBERT architecture to Turkish and to release it together with a
standardised evaluation framework. ModernBERT-TR~\citep{modernberttr} uses the
same architecture. On the multilingual side, mBERT~\citep{bert} and
XLM-R~\citep{xlmr} include Turkish as one language among many, while
mmBERT~\citep{mmbert} scales the ModernBERT architecture to more than 1800
languages.

All of these models monolingual and multilingual alike use pure MLM as
their pretraining objective. This is the point at which MoganBert-TR departs
from them.

\subsection{Pretraining objective and training schedule}
That pure MLM is not the only option for encoder pretraining has recently been
demonstrated under controlled conditions: \citet{clmmlm}, using 38 models
(210M--1B) and more than 15,000 fine-tuning runs, report that under a fixed
compute budget a two-stage \clmmlm{} design consistently outperforms pure MLM.
For the 610M model, the sequence classification average is 87.00 for 100\% MLM,
87.85 for 25\% CLM + 75\% MLM, 87.58 for 50\%/50\%, and 83.58 for 100\% CLM.
The proposed mechanism is that CLM is more data-efficient in early steps and
produces representations less sensitive to the learning rate during
fine-tuning, whereas MLM eventually pulls ahead on representation quality. The
critical implementation detail is that the MLM phase resumes from a CLM
checkpoint that has not yet entered learning-rate decay; in a WSD
schedule~\citep{wsd} the transition is made inside the stable phase.

The final phase of the schedule (annealing/decay) is an independent design
space. mmBERT~\citep{mmbert} shows that data shown in the low-learning-rate
region is disproportionately influential, and that adding low-resource
languages only during this phase can yield large gains. The role of
\emph{context length} within that same phase, however, has not been measured;
Section~\ref{sec:branches} fills this gap.

\subsection{Data pipelines and quality filtering}
The methodological choices behind the pipeline described below follow the
patterns established by CCNet~\citep{ccnet} and
FineWeb/FineWeb2~\citep{fineweb,fineweb2}: direct text extraction from WARC,
fastText-based language identification, and the application of quality and
repetition filters as a sequential chain. FineWeb-Edu~\citep{finewebedu}
demonstrated that a quality classifier learned from LLM labels can be used at
scale. The essential difference in adapting this to Turkish is the absence of an
off-the-shelf Turkish quality classifier and the resulting need to build one
from scratch (Section~\ref{sec:quality}).

\subsection{Embedding models}
The raw representations of MLM-pretrained encoders are anisotropic: all vectors
collect within a narrow cone, and cosine similarity buries the discriminative
signal in noise. This pattern has been characterised in the literature as a
``rogue dimension''~\citep{rogue}. Contrastive fine-tuning and teacher
distillation~\citep{distill} are the established ways of correcting this
geometry; MTEB~\citep{mteb} is used as the evaluation ground.

\section{Pretraining Data}
\label{sec:data}

\subsection{Sources and processing pipeline}
The corpus was compiled from three sources. The first is the Turkish portion of
FineWeb2~\citep{fineweb2}; rather than being used as-is, it was passed again
through the Turkish quality filter described below. The second consists of the
recent months that collection does not cover: Turkish records are first indexed
via each monthly crawl's columnar index, after which only the relevant byte
ranges are fetched with HTTP Range GET the WARC files are not downloaded in
full. Main text is extracted from raw WARC records with \texttt{trafilatura};
the ready-made WET text was deliberately not used because it carries boilerplate
and menu residue (CCNet~\citep{ccnet} and FineWeb~\citep{fineweb} follow the
same route).

After extraction, language verification (fastText \texttt{lid.176.ftz}),
Gopher/C4-style heuristic quality filters, a Turkish boilerplate check, PII
masking and learned quality classification are applied in a single pass; the
output is written directly as clean JSONL. The final step is MinHash-based
fuzzy deduplication~\citep{minhash} (n-gram 5, 14 buckets, 8 hashes per bucket),
which operates incrementally across months: each month is first deduplicated
within itself and then against the accumulated corpus. The processing rate
measured in production is $\sim$430--590 records/second, and the acceptance rate
falls in the 29\%--55\% band depending on the month.

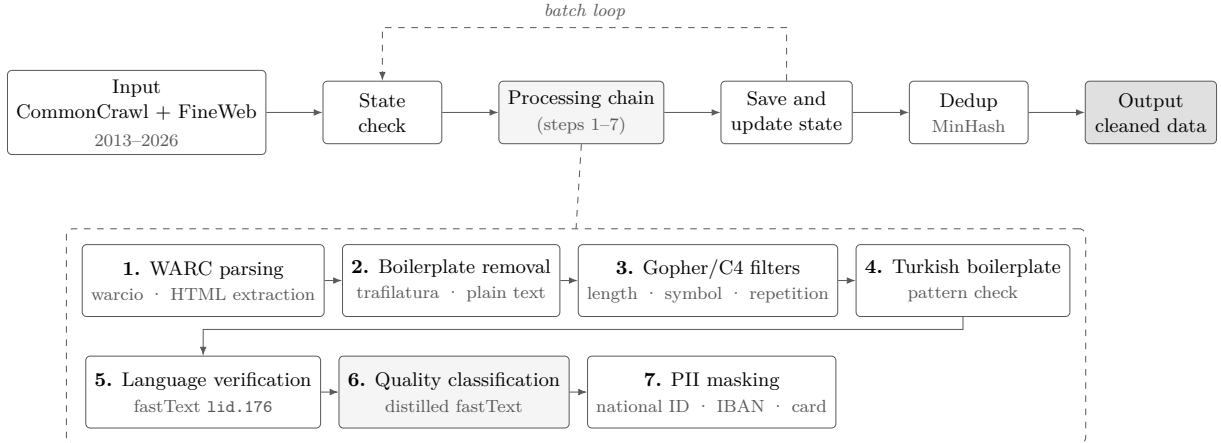
\begin{figure}[H]
\centering
\resizebox{\linewidth}{!}{%
\begin{tikzpicture}[node distance=0pt]

\node[fnode, minimum width=2.5cm, minimum height=1cm] (in)
  {Input\\ CommonCrawl + FineWeb\\[1pt]\textcolor{figline}{\scriptsize 2013--2026}};
\node[fnode, right=0.9cm of in, minimum width=1.9cm, minimum height=1cm] (state)
  {State\\check};
\node[snode, right=0.9cm of state, minimum width=2.6cm, minimum height=1cm] (proc)
  {Processing chain\\\textcolor{figline}{\scriptsize (steps 1--7)}};
\node[fnode, right=0.9cm of proc, minimum width=2.1cm, minimum height=1cm] (save)
  {Save and\\update state};
\node[fnode, right=0.9cm of save, minimum width=1.9cm, minimum height=1cm] (dedup)
  {Dedup\\\textcolor{figline}{\scriptsize MinHash}};
\node[anode, right=0.9cm of dedup, minimum width=2.1cm, minimum height=1cm] (out)
  {Output\\cleaned data};

\draw[farrow] (in)    -- (state);
\draw[farrow] (state) -- (proc);
\draw[farrow] (proc)  -- (save);
\draw[farrow] (save)  -- (dedup);
\draw[farrow] (dedup) -- (out);

\draw[floop] (save.north) -- ++(0,0.85)
  node[midway,right,fsub]{}
  -| (state.north)
  node[pos=0.25,above,flab]{batch loop};

\node[fnode, below=1.6cm of proc, xshift=-6.05cm, minimum width=2.55cm, minimum height=1.15cm, anchor=north] (s1)
  {\textbf{1.} WARC parsing\\\textcolor{figline}{\scriptsize warcio · HTML extraction}};
\node[fnode, right=0.28cm of s1, minimum width=2.55cm, minimum height=1.15cm] (s2)
  {\textbf{2.} Boilerplate removal\\\textcolor{figline}{\scriptsize trafilatura · plain text}};
\node[fnode, right=0.28cm of s2, minimum width=2.55cm, minimum height=1.15cm] (s3)
  {\textbf{3.} Gopher/C4 filters\\\textcolor{figline}{\scriptsize length · symbol · repetition}};
\node[fnode, right=0.28cm of s3, minimum width=2.55cm, minimum height=1.15cm] (s4)
  {\textbf{4.} Turkish boilerplate\\\textcolor{figline}{\scriptsize pattern check}};

\node[fnode, below=0.62cm of s1, minimum width=2.55cm, minimum height=1.15cm] (s5)
  {\textbf{5.} Language verification\\\textcolor{figline}{\scriptsize fastText \texttt{lid.176}}};
\node[snode, right=0.28cm of s5, minimum width=2.55cm, minimum height=1.15cm] (s6)
  {\textbf{6.} Quality classification\\\textcolor{figline}{\scriptsize distilled fastText}};
\node[fnode, right=0.28cm of s6, minimum width=2.55cm, minimum height=1.15cm] (s7)
  {\textbf{7.} PII masking\\\textcolor{figline}{\scriptsize national ID · IBAN · card}};

\draw[farrow] (s1) -- (s2);  \draw[farrow] (s2) -- (s3);  \draw[farrow] (s3) -- (s4);
\draw[farrow] (s5) -- (s6);  \draw[farrow] (s6) -- (s7);
\draw[farrow] (s4.south) -- ++(0,-0.20) -- ($(s1.south)+(0,-0.20)$) -- ($(s5.north)+(0,0)$);

\begin{scope}[on background layer]
  \node[gnode, fit=(s1)(s4)(s5)(s7)] (box) {};
\end{scope}
\draw[draw=figline, line width=0.4pt, dashed] (proc.south) -- (box.north);

\end{tikzpicture}}
\caption{The data pipeline: from FineWeb2 and Common Crawl input through a
single-pass processing chain to the deduplicated output. Quality classification
(step~6) is not a separate pass but part of the same process.}
\label{fig:pipeline}
\end{figure}

\subsection{The quality classifier: from teacher to label generator}
\label{sec:quality}
The only Turkish-specific gap in the pipeline was the quality classifier:
heuristic filters do not remove text that is grammatically valid but worthless
(gambling, SEO, advertising), and no off-the-shelf Turkish quality classifier
exists.

A three-stage chain was built. A fastText classifier trained with off-the-shelf
word vectors failed by discarding legitimate short text as well; retrained on
$\sim$6,000 documents scored by an LLM against a 0--5 rubric (following the
FineWeb-Edu~\citep{finewebedu} pattern), fastText remained at $\sim$54\%
validation accuracy. A Turkish BERT fine-tuned on the same labels reached 70.7\%
exact and 93.9\% $\pm$1 accuracy, but at $\sim$96.5 documents/second it was
unsuitable for production.

The solution was to use BERT not as the production classifier but as a
\emph{label generator}: a binary fastText model was trained from scratch on the
$\sim$480,000 keep/discard decisions accumulated across the processed chunks. On
a chunk the teacher had never seen, the distilled model gave \textbf{94.4\%
agreement} and $\sim$8,880 documents/second a
\textbf{$\sim$90$\times$} speed-up with no loss of accuracy~\citep{distill}.
This pattern transfers directly to any language lacking an off-the-shelf
quality classifier.

\subsection{Printed and institutional sources}
Alongside the web branch, a second track was added for domain-dense,
editorially reviewed content: text extraction from books, theses and
academic/legal publications. The decisive constraint here is not transcription
accuracy but \emph{structural classification} the cover page, copyright
notice, preface and table of contents are not the content of a book and must
not enter the corpus. Classical OCR tools cannot make this distinction because
they decide on headings from font size and position; a vision-language model
that solves both layers in a single request was therefore used instead. As the
output format, line-tagged plain text was preferred over JSON; JSON produced 91
parse errors over 448 pages because of quotation marks and special characters in
Turkish text. In post-processing, front-matter blocks are discarded, identical
section headings reopened at page boundaries are merged, and text is chunked
with a target of $\sim$6,000 characters. Details are given in
Appendix~\ref{app:data}.

\section{Tokenizer}
\label{sec:tokenizer}

The tokenizer is a 50,048-token SentencePiece Unigram
model~\citep{sentencepiece}. This section presents three design decisions and
the measured efficiency; implementation details such as the normalization
pipeline and code sensitivity are given in Appendix~\ref{app:tokenizer}.

\subsection{Vocabulary size}
Three vocabulary sizes were trained and measured on TR-MMLU (6,200 questions,
1,743,375 characters). The linguistic metrics (TR\% = the proportion of valid
Turkish unique tokens, Pure\% = the proportion of atomic tokens) follow the
definitions of \citet{trtok}, which reports a correlation of $r=0.90$ between
TR\% and MMLU score. Dilbaz (60,000+ roots, FST + rule engine) was used for
morphological analysis.

\begin{table}[H]
\centering
\footnotesize
\caption{Tokenizer efficiency on TR-MMLU.}
\label{tab:tok-trmmlu}
\begin{tabular}{lrrrrr}
\toprule
\textbf{Tokenizer} & \textbf{Vocab} & \textbf{Fertility $\downarrow$} &
\textbf{Chars/Token $\uparrow$} & \textbf{TR\% $\uparrow$} & \textbf{Pure\% $\uparrow$} \\
\midrule
\ours Mogan-64K & 64,000 & \textbf{1.5268} & \textbf{5.0245} & \textbf{84.71} & 40.83 \\
\ours Mogan-50K & 50,000 & 1.5682 & 4.8918 & 84.42 & 42.67 \\
Mürşit & 59,008 & 1.6326 & 4.6989 & 77.00 & 35.81 \\
MBERT-tr & 32,000 & 1.7037 & 4.5029 & 78.73 & 35.65 \\
BERTurk & 32,000 & 1.8008 & 4.2600 & 81.06 & \textbf{43.22} \\
TabiBERT & 50,176 & 1.8501 & 4.1464 & 59.84 & 30.46 \\
XLM-R & 250,002 & 1.9440 & 3.9462 & 56.09 & 41.78 \\
\bottomrule
\end{tabular}
\end{table}

The 64K vocabulary beats 50K on every efficiency metric (fertility 1.527 vs
1.568), but this gain comes at the cost of embedding parameters: in a model
with hidden size 768, 14,000 additional tokens amount to $\approx$ 10.7M
additional parameters roughly 7\% of the parameter budget of a 150M
foundation model. Since a 2.7\% fertility difference does not justify that cost,
the 50K base was chosen. The final size is \textbf{50,048}: the multiple of 64
closest to 50,000; GPU tensor cores operate more efficiently at sizes that are
multiples of 64.

\begin{figure}[H]
\centering
\includegraphics[width=0.74\linewidth]{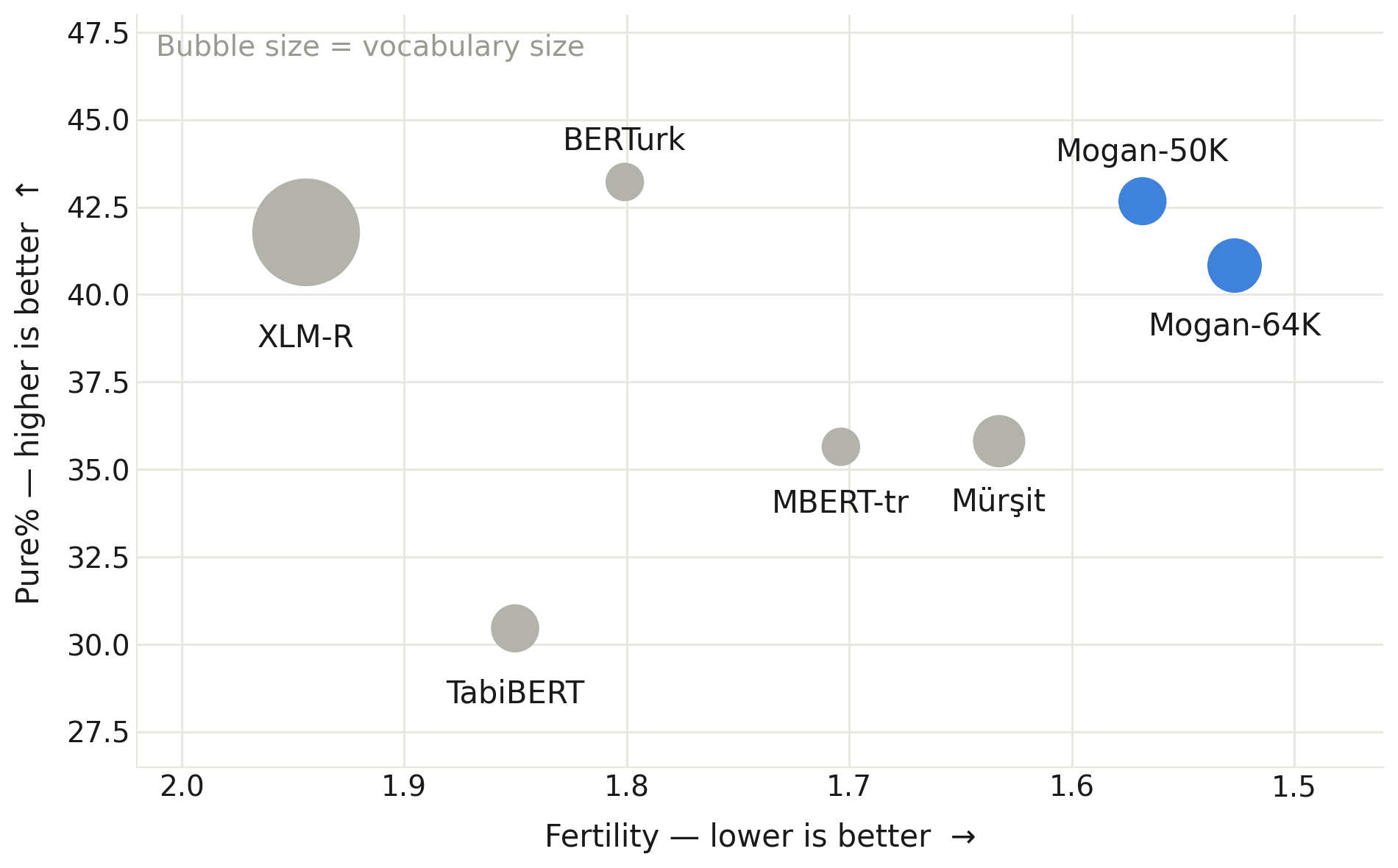}
\caption{Fertility ($x$, lower is better) vs Pure\% ($y$, higher is better);
bubble size indicates vocabulary size.}
\end{figure}

\subsection{Validation on independent test sets}
In addition to TR-MMLU, the tokenizer was tested on two further sources
entirely independent of the data collected for this project: Turkish Wikipedia
(\texttt{20231101.tr}, 30,000 documents) and FLORES-200~\citep{flores}
(\texttt{tur\_Latn}, dev+devtest, 2,009 sentences), the academic standard for
multilingual evaluation. In both tests the same document set was applied to all
tokenizers under the same seed (42).

\begin{table}[H]
\centering
\footnotesize
\caption{Tokenizer efficiency on independent test sets.}
\label{tab:tok-indep}
\begin{tabular}{llrrrr}
\toprule
\textbf{Test set} & \textbf{Tokenizer} & \textbf{Vocab} &
\textbf{Chars/Token $\uparrow$} & \textbf{Fertility $\downarrow$} & \textbf{UNK} \\
\midrule
\ours Wikipedia & \textbf{Mogan-64K} & 64,000 & \textbf{4.705} & \textbf{1.640} & 0.00\% \\
Wikipedia & BERTurk & 32,000 & 4.391 & 1.757 & 0.04\% \\
Wikipedia & Kumru-2B & 50,176 & 4.010 & 1.924 & 0.00\% \\
Wikipedia & TabiBERT & 50,176 & 4.010 & 1.924 & 0.00\% \\
\midrule
\ours FLORES-200 & \textbf{Mogan-64K} & 64,000 & \textbf{5.277} & \textbf{1.460} & 0.00\% \\
FLORES-200 & BERTurk & 32,000 & 4.932 & 1.563 & 0.002\% \\
FLORES-200 & TabiBERT & 50,176 & 4.900 & 1.573 & 0.00\% \\
FLORES-200 & Kumru-2B & 50,176 & 4.899 & 1.573 & 0.00\% \\
\bottomrule
\end{tabular}
\end{table}

On both independent test sets the Mogan tokenizer outperforms all compared
tokenizers on compression as well as on fertility.

\subsection{Code sensitivity and special tokens}
\label{sec:code-tok}
In code data, indentation is part of the syntax; because standard normalization
removes it, different programs can collapse onto the same token sequence. With
an indentation-preserving pipeline and the
\texttt{allow\_whitespace\_only\_pieces} setting, the model learned 11
multi-space tokens ranging from 3 to 16 spaces.

\begin{table}[H]
\centering
\footnotesize
\caption{Tokenizer comparison on a code corpus (18 languages, 540 files).
Lossless roundtrip is the rate at which a model can reproduce its own output.}
\label{tab:tok-code}
\begin{tabular}{lrr}
\toprule
\textbf{Tokenizer} & \textbf{Code compression} & \textbf{Lossless roundtrip} \\
\midrule
\ours \textbf{Mogan (final)} & \textbf{2.790} & \textbf{100\%} \\
Mürşit & 2.454 & 96\% \\
TabiBERT & 2.196 & 49\% \\
BERTurk & 2.184 & 0\% \\
\bottomrule
\end{tabular}
\end{table}

In the placement of special tokens the single critical decision is keeping
\texttt{[MASK]} in the \texttt{control\_symbols} list: web data contains
documents that write \texttt{[MASK]} literally, and if the other list is used
that text is converted into the actual mask identifier, injecting spurious masks
into the training data. The final layout and the rejected alternatives are given
in Appendix~\ref{app:tokenizer}.

\section{Model and Pretraining}
\label{sec:model}

\subsection{Architecture}
The model follows the ModernBERT-base configuration: 22 layers, hidden size
768, alternating local/global attention (local window 128), RoPE and GLU; 149.4M
parameters in total. Since HuggingFace's ModernBERT implementation is
bidirectional only, the attention path was rewritten to support the CLM phase.
This rewrite involves four pitfalls capable of silently producing an incorrect
model (causal local window, boundary-aware packing, position-id resetting,
document-boundary masking in the loss); these are documented together with the
memory and numerical decisions in Appendix~\ref{app:impl}.

\subsection{Training objective: the \clmmlm{} curriculum}
\label{sec:objective}
The first portion of the run is trained with the causal language modelling
(CLM) objective and the remainder with masked language modelling (MLM). The
transition is made inside the stable phase of a WSD schedule~\citep{wsd} and
without touching the learning rate; decay comes only at the very end. At the
same step, the attention window is switched from causal to bidirectional mode.

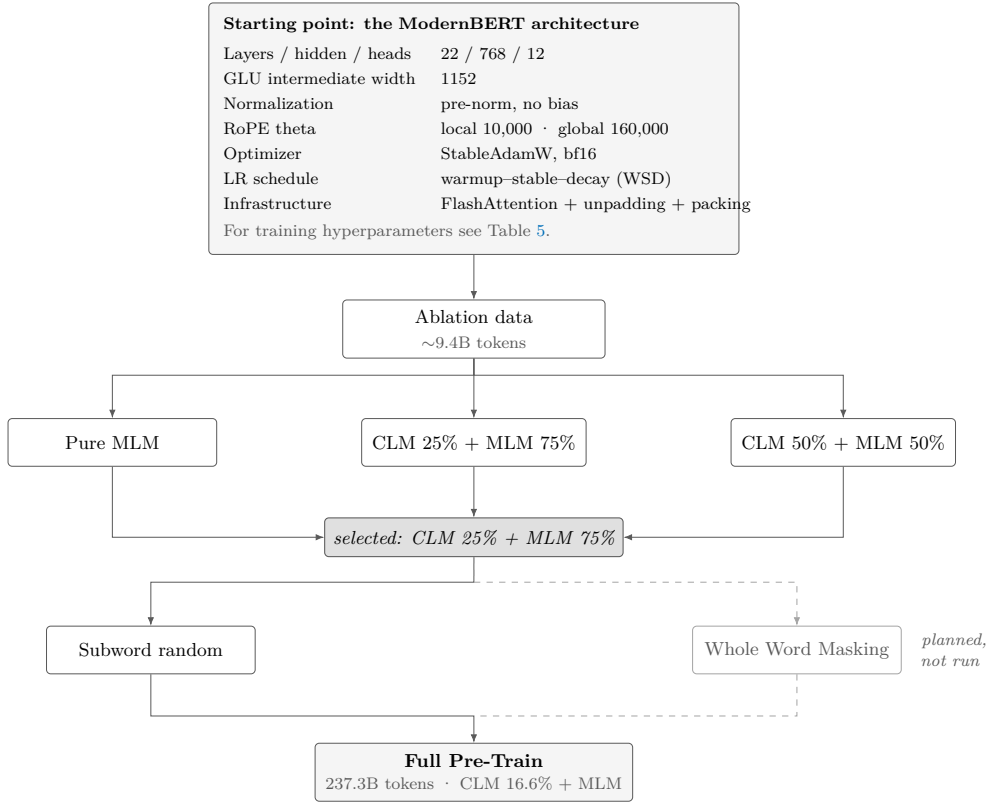
\begin{figure}[H]
\centering
\resizebox{0.88\linewidth}{!}{%
\begin{tikzpicture}[node distance=0.55cm]

\tikzset{
  fb/.style   = {draw=figline, line width=0.4pt, rounded corners=2pt,
                  fill=white, align=center, inner sep=5pt, font=\footnotesize,
                  minimum width=3.5cm, minimum height=0.8cm},
  pick/.style  = {draw=figline, line width=0.5pt, rounded corners=2pt,
                  fill=figaccent, align=center, inner sep=4pt,
                  font=\footnotesize\itshape, minimum height=0.65cm},
  final/.style = {fb, fill=figfill, font=\footnotesize\bfseries,
                  minimum width=4cm, minimum height=0.9cm},
  ghost/.style = {fb, draw=figline!55, text=figline},
}

\node[draw=figline, line width=0.4pt, rounded corners=2pt, fill=figfill,
      align=left, inner sep=7pt, font=\scriptsize, text width=8.4cm] (recipe)
{\textbf{Starting point: the ModernBERT architecture}\\[3pt]
\begin{tabular}{@{}ll@{}}
Layers / hidden / heads & 22 / 768 / 12 \\
GLU intermediate width & 1152 \\
Normalization & pre-norm, no bias \\
RoPE theta & local 10,000 · global 160,000 \\
Optimizer & StableAdamW, bf16 \\
LR schedule & warmup--stable--decay (WSD) \\
Infrastructure & FlashAttention + unpadding + packing \\
\end{tabular}\\[3pt]
\textcolor{figline}{\scriptsize For training hyperparameters see Table~\ref{tab:config}.}};

\node[fb, below=0.75cm of recipe, minimum width=4.4cm] (data)
  {Ablation data\\\textcolor{figline}{\scriptsize $\sim$9.4B tokens}};
\draw[farrow] (recipe) -- (data);

\node[fb, below left=1.0cm and 2.1cm of data] (mlm)   {Pure MLM};
\node[fb, below=1.0cm of data]                 (c25)  {CLM 25\% + MLM 75\%};
\node[fb, below right=1.0cm and 2.1cm of data] (c50)  {CLM 50\% + MLM 50\%};

\draw[farrow] (data.south) -- ++(0,-0.28) -| (mlm.north);
\draw[farrow] (data.south) -- ++(0,-0.28) -| (c25.north);
\draw[farrow] (data.south) -- ++(0,-0.28) -| (c50.north);

\node[pick, below=0.85cm of c25, minimum width=4.6cm] (win1)
  {selected: CLM 25\% + MLM 75\%};
\draw[farrow] (mlm.south) |- (win1.west);
\draw[farrow] (c50.south) |- (win1.east);
\draw[farrow] (c25) -- (win1);

\node[fb, below left=1.15cm and 1.15cm of win1]  (sub) {Subword random};
\node[ghost, below right=1.15cm and 1.15cm of win1] (wwm) {Whole Word Masking};
\draw[farrow] (win1.south) -- ++(0,-0.42) -| (sub.north);
\draw[farrow, draw=figline!55, dashed] (win1.south) -- ++(0,-0.42) -| (wwm.north);
\node[flab, right=0.2cm of wwm, text width=2.4cm, align=left]
  {planned,\\not run};

\coordinate (m2) at ($(sub.south)!0.5!(wwm.south)$);
\node[final, below=1.15cm of m2] (full)
  {Full Pre-Train\\[-1pt]\textcolor{figline}{\scriptsize\mdseries 237.3B tokens · CLM 16.6\% + MLM}};
\draw[farrow] (sub.south) |- ([yshift=0.45cm]full.north) -- (full.north);
\draw[draw=figline!55, line width=0.4pt, dashed] (wwm.south) |- ([yshift=0.45cm]full.north);

\end{tikzpicture}}
\caption{Decision flow for the \clmmlm{} curriculum. The first ablation
(§\ref{sec:ablation}) selects the training objective; the masking ablation was
designed but not run on the basis of corpus statistics (dashed branch,
§\ref{sec:masking}). Training hyperparameters are given in
Table~\ref{tab:config}.}
\label{fig:curriculum}
\end{figure}

\subsection{Ablation: \clmmlm{} vs.\ pure MLM}
\label{sec:ablation}
The claim reported at English scale by \citet{clmmlm} was tested on the
MoganBert architecture, the MoganBert tokenizer and the Turkish corpus
described above.

\paragraph{Setup.} Both runs were trained with the same data pipeline, the same
architecture and the same total number of steps (10,000): \texttt{mlm\_only}
runs 10,000 steps of pure MLM; \texttt{clm25\_mlm75} uses CLM for the first 25\%
($\sim$2,500 steps) and MLM for the remaining 75\%. The ablations were run on a
$\sim$9.4-billion-token subset of the full corpus. The evaluation protocol is
identical for both checkpoints (seed 1234, mask rate 0.30, \texttt{max\_len}
512).

\paragraph{An important asymmetry.} At equal total steps, \texttt{mlm\_only}
has seen 33\% more MLM steps than \texttt{clm25\_mlm75} (10,000 vs 7,500). The
differences below are obtained in favour of the curriculum \emph{despite} this
handicap.

\begin{table}[H]
\centering
\footnotesize
\caption{Ablation results. The probes are linear classifiers trained on frozen
representations; MS MARCO was run without contrastive training, on the raw
pretraining representations. In the embedding geometry block, \emph{lower} is
better on the first two rows.}
\label{tab:ablation}
\begin{tabular}{lrrr}
\toprule
\textbf{Metric} & \textbf{MLM-only} & \textbf{CLM25+MLM75} & \textbf{$\Delta$} \\
\midrule
\multicolumn{4}{l}{\emph{Linear probe (accuracy \%)}} \\
\quad POS (BOUN), 15 classes & \textbf{87.62} & 87.42 & $-$0.20 \\
\quad News category, 5 classes & 97.39 & 97.39 & 0.00 \\
\quad NLI, 3 classes & 48.53 & \textbf{50.22} & +1.69 \\
\quad Product review, 2 classes & 80.75 & \textbf{81.80} & +1.05 \\
\quad \textbf{Average} & 78.57 & \textbf{79.21} & \textbf{+0.64} \\
\midrule
\multicolumn{4}{l}{\emph{Turkish MS MARCO (\%), mean pooling}} \\
\quad R@1 & 2.96 & \textbf{10.86} & +7.90 \;($\times$3.67) \\
\quad R@10 & 9.92 & \textbf{26.34} & +16.42 \;($\times$2.66) \\
\quad MRR & 5.59 & \textbf{16.23} & +10.64 \;($\times$2.90) \\
\midrule
\multicolumn{4}{l}{\emph{Embedding geometry}} \\
\quad Variance share of 1st component & 0.2810 & \textbf{0.1187} & $-$0.1623 \\
\quad \texttt{cos\_raw} & 0.9731 & \textbf{0.9595} & $-$0.0136 \\
\quad STS Spearman & 0.4833 & \textbf{0.5026} & +0.0193 \\
\bottomrule
\end{tabular}
\end{table}

\paragraph{Finding.} On the probe side the two models are practically
\emph{tied}; the +0.64-point difference in the average was measured with a
single seed and cannot be separated from within-run noise. On POS the
difference is 0.20 points, and on news category the two models give identical
accuracy over 230 examples that task is saturated. Meaningful signal
appears only on NLI and product review, both of which require sentence-level
semantic representation. On the MS MARCO side, by contrast, the separation is
qualitatively different: the curriculum model leads by 2.7--3.7$\times$ on every
metric, which cannot be explained by single-seed noise.

\paragraph{Mechanism: embedding geometry.} That the probes are tied while
retrieval differs threefold is no contradiction; the two metrics are sensitive
to different things. A probe is a learned linear transformation and can rescale
or ignore dominant directions that carry no information; cosine retrieval
cannot, as it depends directly on the raw geometry. The critical row is the
variance share of the first component: in the pure-MLM model a single direction
absorbs 28.1\% of the variance, against 11.9\% in the curriculum model. Known in
the literature as a ``rogue dimension'', this pattern~\citep{rogue} compresses
all vectors towards a common direction, and because cosine similarity cannot
remove that common component, the discriminative signal is buried in noise. The
consistent difference on STS which is likewise cosine-based is an
independent confirmation of the same mechanism.

\paragraph{Decision.} Full pretraining was carried out with the \clmmlm{}
curriculum: parity at worst on the probe metrics, against a clear and
unidirectional advantage on all cosine-based tasks and with 33\% fewer MLM
steps at that.

\subsection{Masking strategy}
\label{sec:masking}
Which masking strategy to use in the MLM phase (subword random, whole word
masking, or morphology-aware) was evaluated from corpus statistics without
spending a training run. Three thousand documents per domain were sampled from
five domains ($\sim$15.9M words, 698,865 unique types). The token-weighted rate
of words split into $\geq$2 pieces is 41.1\% and the type-weighted rate is
95.0\%; the token shares of root and suffix candidates are 76.6\% / 23.4\%.

This gap is methodologically decisive: 95\% of word types are multi-piece, but
these are rare words; only 41\% of the actual training token stream comes from
multi-piece words. That number is directly the upper bound on WWM on 59\% of
the token stream, WWM and token-level random masking are already identical. The
share that suffix-targeted masking could touch is 23.4\%. \textbf{Decision:}
token-level random masking was used in pretraining; a ceiling of 41\% did not
justify a separate ablation.

\subsection{Full pretraining configuration}
\begin{table}[H]
\centering
\footnotesize
\caption{Training configuration.}
\label{tab:config}
\begin{tabular}{ll}
\toprule
\textbf{Item} & \textbf{Value} \\
\midrule
Parameter count & 149.4M (22 layers, hidden 768, \texttt{local\_attention}=128) \\
Vocabulary & 50,048 \\
Hardware & 4$\times$ H100 (DDP) \\
Attention & FlashAttention-3~\citep{flashattn}, varlen (\texttt{cu\_seqlens}) \\
Precision & bf16 \\
Optimizer & StableAdamW, betas (0.90, 0.98), eps 1e-6, weight decay 1e-5 \\
Peak LR & 8e-4 (WSD; warmup 1,431 steps, inside the CLM phase) \\
Global batch & 2048 blocks $\times$ 1024 tokens = 2,097,152 tokens/step \\
Gradient accumulation & 16 \\
Gradient clipping & 1.0 \\
MFU & 0.248 \\
Throughput & 836K tokens/second (4 GPUs in total) \\
\bottomrule
\end{tabular}
\end{table}

\subsection{Phase plan and branched annealing}
\label{sec:phaseplan}
The main run is a single stretch; the annealing phase is \emph{split into two
branches} after a shared prefix.

\begin{table}[H]
\centering
\footnotesize
\caption{Phase plan. 237.3B tokens in total for each branch.}
\label{tab:phases}
\begin{tabular}{lrlrrl}
\toprule
\textbf{Phase} & \textbf{Tokens} & \textbf{Steps} & \textbf{Ctx} & \textbf{Mask} & \textbf{LR} \\
\midrule
CLM & 36.0B & 1 -- 17,167 & 1024 & --- & warmup + stable \\
MLM & 180.7B & 17,168 -- 103,342 & 1024 & 30\% & 8e-4 stable \\
\midrule
\multicolumn{6}{l}{\emph{Annealing --- shared prefix}} \\
Anneal (shared) & 10.5B & 1 -- 5,000 & 8192 & 10\% & decay, upper half \\
\midrule
\multicolumn{6}{l}{\emph{Annealing --- two branches}} \\
Branch A: \texttt{anneal} & 10.1B & 5,001 -- 9,810 & 8192 & 10\% & decay, lower half \\
\ours Branch B: \texttt{anneal1k} & 10.1B & 5,001 -- 9,810 & \textbf{1024} & 10\% & decay, lower half \\
\bottomrule
\end{tabular}
\end{table}

Three things change simultaneously in the annealing phase. \textbf{(i)} Context
is extended from 1024 to 8192; the global RoPE theta is scaled from 10,000 to
160,000, while the local window theta is left at 10,000 there is nothing to
extend within a 129-token window. \textbf{(ii)} The mask rate is lowered from
30\% to 10\%; a lower mask rate produces a cleaner language modelling signal in
this phase. \textbf{(iii)} The data mixture is shifted in favour of
high-quality Turkish, because data shown in the low-LR region leaves a more
lasting imprint on the model~\citep{mmbert}.

\paragraph{Rationale for branching.} The alternative was to merge two different
regimes with a \emph{model soup}~\citep{soup}. Branching is cheaper: the shared
prefix is run once and only the final 4,810 steps are computed twice a total
additional cost of $\sim$4.3\%. A soup, by contrast, requires a separate full
decay run for each regime. Both were produced and compared
(Section~\ref{sec:branches}); before the soup, the relative distance in weight
space was measured ($|\text{main} - \text{anneal}| / |\text{main}| = 0.1011$
the same basin, so averaging is meaningful).

The LR decay curve (1-sqrt, 8e-4 $\rightarrow$ 8.36e-6) is computed over the
entire annealing run (9,810 steps) in both branches; branching changes only the
\emph{context length} of the final 4,810 steps. That is the only variable
between the two branches.

\subsection{Data mixture and held-out set}
The mixture was applied not through probabilistic sampling during training but
through a precomputed recipe file; this makes the ratios \emph{exact} rather
than statistical and makes resuming fully deterministic. The unique corpus
comprises 107.96 billion tokens; the main run is two epochs over it.

\begin{table}[H]
\centering
\footnotesize
\caption{Realised group ratios in the main run. Within-step deviation is 0.055
points.}
\label{tab:mixture}
\begin{tabular}{lrrr}
\toprule
\textbf{Group} & \textbf{Target} & \textbf{Realised} & \textbf{$\Delta$} \\
\midrule
CommonCrawl & 52.6\% & 52.60\% & 0.00 \\
High-quality Turkish & 20.7\% & 20.83\% & +0.13 \\
\texttt{fineweb\_edu\_eng} & 13.9\% & 13.84\% & $-$0.06 \\
\texttt{coding\_gold} & 10.0\% & 9.97\% & $-$0.03 \\
\texttt{eng\_math\_gold} & 2.8\% & 2.77\% & $-$0.03 \\
\bottomrule
\end{tabular}
\end{table}

The language distribution is $\sim$73\% Turkish, $\sim$17\% English and
$\sim$10\% code. The code share was deliberately kept high so that the
tokenizer's lossless-roundtrip advantage on code
(Section~\ref{sec:code-tok}) would be reflected in the model. For the two
sources whose pool exceeds the amount to be used, a \emph{different} slice was
sampled in each epoch, so that twice as much unique data was seen at the same
cost.

The evaluation set was separated from the training corpus before packing
through a MinHash-based split: 6,202 documents, 136--200 per source. Candidates
with a match in the training data were discarded.

\paragraph{The CLM ratio.} The CLM share is 16.6\% of the main run (36.0B /
216.7B), whereas the ratio validated in Section~\ref{sec:ablation} was 25\%. The
reason for the deviation is a variable the ablation does not cover: the ablation
runs were measured in a regime in which not even a single epoch was completed,
whereas full pretraining was designed over two epochs. In a two-epoch design,
the set seen during the CLM phase receives one MLM pass while the remainder
receives two; the magnitude of this asymmetry has not been measured, and the
ratio was deliberately kept low under that uncertainty. \textbf{This value has
not been validated by ablation.}

\section{Evaluation}
\label{sec:eval}

\subsection{Protocol}
Evaluation uses two independent Turkish benchmarks. TrGLUE~\citep{trglue} is the
primary benchmark: it is sentence-level, has a fixed five-seed protocol, and
therefore supports the paired-seed comparisons on which the annealing analysis
of Section~\ref{sec:branches} rests. TabiBench~\citep{tabibert} is the secondary
benchmark: it is broader (28 datasets, eight categories, including retrieval and
code) but single-seed, so it measures coverage rather than statistical
separation.

\paragraph{TrGLUE.}
The TrGLUE evaluation was carried out entirely with the official
\texttt{run\_trglue.py} script; no metric was reimplemented. Hyperparameters
were taken from the repository README: lr 3e-5 and batch 16 for
\texttt{rte/stsb/mrpc}, lr 2e-5 and batch 128 for the remaining tasks;
\texttt{max\_seq\_length} 128, 5 epochs, full fine-tuning. The seeds are the
five official values given in the README (1, 4, 21, 40, 124), and every reported
number is the average over those five. Since the repository provides no
aggregation code reducing the eight tasks to a single score, the distinction is
stated explicitly: for \texttt{mrpc}, \texttt{stsb} and \texttt{qqp} the
script's own \texttt{eval\_combined\_score} is used; \texttt{cola},
\texttt{sst2}, \texttt{qnli} and \texttt{rte} are reported with a single
metric; for \texttt{mnli} the matched/mismatched average is taken. For all
models, \texttt{classifier\_pooling} was set to \texttt{mean}.

\paragraph{TabiBench.} The evaluation follows the reference procedure
of~\citet{tabibert} without modification. Each of the 28 tasks is run in two
stages. \emph{Stage 1 --- hyperparameter search:} 16 combinations over learning
rate (5e-6, 1e-5, 2e-5, 3e-5), weight decay (1e-5, 1e-6) and batch size (16, 32)
are tried on the \emph{validation} split only, each for at most 10 epochs with
early stopping; the winning configuration is selected on validation performance.
\emph{Stage 2 --- test measurement:} a single run with the winning configuration
is performed and only its \emph{test} score is reported. The test split is not
used at any point during hyperparameter selection. All 28 tasks go through this
same procedure with identical early-stopping rules; there is no methodological
difference between tasks. The learning-rate schedule is 6\% warmup followed by
linear decay to 0.02$\times$ the peak value, with mixed precision and seed 25.

Scores are aggregated as in the reference work: within a category, a
\emph{weighted average} proportional to test-set size; from categories to the
overall score, a \emph{macro average} with all eight categories weighted
equally. The aggregation was verified against the reference paper's own numbers:
recomputing the weighted averages from the per-task scores and test sizes in its
appendix reproduces the published \emph{Weighted Avg} column on 40 of 40 rows to
within 0.007 points, and the macro average of the categories reproduces the
published \emph{Total Avg} exactly (TabiBERT: $620.65 / 8 = 77.58$). Test set
sizes were taken from the same appendix and agree with the Hugging Face dataset
cards on 28 of 28 tasks.

\paragraph{Calibration check.} The correctness of the protocol was tested on
two independent models with published numbers: BERTurk 79.87 (published 80.07,
$-$0.20) and ModernBERT-TR 77.64 (published 78.23, $-$0.59). That BERTurk holds
within $\pm$0.2 indicates the protocol is sound; the deviation on ModernBERT-TR
is most likely due to the published number having been obtained under that
work's own settings rather than the TrGLUE protocol.

\subsection{Branched annealing: two branches and a model soup}
\label{sec:branches}
The three candidates produced in Section~\ref{sec:phaseplan} decay at 8192
(\texttt{anneal}), decay at 1024 (\texttt{anneal1k}), and the weight average of
the main run and anneal (\texttt{soup}) were compared under the same
protocol.

\paragraph{The effect of annealing itself.} Held-out MLM loss (block 8192, 31
blocks, 254K tokens; mask rate fixed at 30\% for measurement) falls monotonically
throughout annealing: starting from the anchor (the final checkpoint of the main
run) at 1.8026, it descends to 1.7072, 1.6801, 1.6607, 1.6474 and 1.6379
($-$0.165). There is no plateau. The critical finding is that even though
training was carried out with a 10\% mask rate, improvement is continuous on the
harder 30\% task as well lowering the mask rate did not damage performance
on the harder task.

\paragraph{Fill-mask probe.} Over 50 sentences and 39 single-token gold labels,
the soup beats both of its parents (top1 71.8\% vs 69.2\% / 64.1\%; MRR 0.787;
mean $\log P$ $-$1.27) the classic soup result. \textbf{This gain, however,
did not carry over downstream.}

\begin{table}[H]
\centering
\footnotesize
\caption{TrGLUE average of the three candidates (five official seeds).}
\label{tab:candidates}
\begin{tabular}{lrr}
\toprule
\textbf{Candidate} & \textbf{TrGLUE AVG} & \textbf{Rank (6 models)} \\
\midrule
\ours \textbf{\texttt{anneal1k}} (decay at 1024) & \textbf{78.41 $\pm$ 0.32} & \textbf{2} \\
\texttt{anneal} (decay at 8192) & 77.92 $\pm$ 0.48 & 3 \\
\texttt{soup} & 77.66 $\pm$ 0.37 & 5 \\
\bottomrule
\end{tabular}
\end{table}

\paragraph{Paired comparison.} When the two branches are compared on the same
seeds, seed-induced noise drops: the differences are +0.79, +0.36, +0.65, +0.13
and +0.52 respectively, averaging $+0.49 \pm 0.26$. Here $t = 4.28$ (df$=4$),
$p \approx 0.013$, and \textbf{all five of the five seeds are positive.} Running
the final portion of decay at 1024 context yielded a statistically significant
improvement.

\textbf{Conclusion:} \texttt{anneal1k} was selected as the final model. The
branching approach is both cheaper ($\sim$4.3\% additional cost) and more
successful (+0.75 TrGLUE points) than the soup. The soup case also concretely
demonstrates the risk of selecting a model on the basis of a small-sample
internal metric.

\subsection{TrGLUE}
\begin{table}[H]
\centering
\footnotesize
\setlength{\tabcolsep}{4pt}
\caption{TrGLUE eight tasks, average $\pm$ standard deviation over the five
official seeds. Bold: best in row.}
\label{tab:trglue}
\begin{tabular}{lrrrrr}
\toprule
\textbf{Task} & \textbf{BERTurk} & \textbf{MBERT-TR} & \textbf{TabiBERT} &
\textbf{Mogan-anneal} & \textbf{Mogan-anneal1k} \\
\midrule
cola & \textbf{41.71}\tiny{$\pm$1.87} & 26.62\tiny{$\pm$2.36} & 31.89\tiny{$\pm$1.76} & 35.92\tiny{$\pm$2.29} & 37.63\tiny{$\pm$1.97} \\
sst2 & \textbf{87.62}\tiny{$\pm$0.19} & 86.47\tiny{$\pm$0.44} & 85.72\tiny{$\pm$0.27} & 85.60\tiny{$\pm$0.13} & 85.83\tiny{$\pm$0.11} \\
mrpc & \textbf{71.00}\tiny{$\pm$0.94} & 70.45\tiny{$\pm$0.63} & 68.91\tiny{$\pm$0.77} & 68.39\tiny{$\pm$1.55} & 69.18\tiny{$\pm$0.56} \\
stsb & \textbf{72.08}\tiny{$\pm$0.60} & 70.51\tiny{$\pm$0.98} & 69.18\tiny{$\pm$5.38} & 68.96\tiny{$\pm$1.28} & 69.59\tiny{$\pm$1.42} \\
qqp & 95.04\tiny{$\pm$0.07} & \textbf{95.17}\tiny{$\pm$0.03} & 95.08\tiny{$\pm$0.08} & 94.46\tiny{$\pm$0.06} & 94.54\tiny{$\pm$0.08} \\
mnli & \textbf{89.11}\tiny{$\pm$0.11} & 88.13\tiny{$\pm$0.30} & 88.00\tiny{$\pm$0.21} & 88.10\tiny{$\pm$0.14} & 88.24\tiny{$\pm$0.19} \\
qnli & \textbf{90.48}\tiny{$\pm$0.17} & 89.94\tiny{$\pm$0.18} & 89.89\tiny{$\pm$0.13} & 89.64\tiny{$\pm$0.17} & 89.87\tiny{$\pm$0.16} \\
rte & 91.94\tiny{$\pm$0.50} & 93.84\tiny{$\pm$0.57} & \textbf{93.94}\tiny{$\pm$0.34} & 92.30\tiny{$\pm$0.75} & 92.42\tiny{$\pm$0.40} \\
\midrule
\ours \textbf{AVG} & \textbf{79.87}\tiny{$\pm$0.23} & 77.64\tiny{$\pm$0.37} & 77.83\tiny{$\pm$0.57} & 77.92\tiny{$\pm$0.48} & \textbf{78.41}\tiny{$\pm$0.32} \\
\bottomrule
\end{tabular}
\end{table}

MoganBert-TR (\texttt{anneal1k}) is the best of the Turkish ModernBERT models
compared (78.41 vs TabiBERT 77.83, ModernBERT-TR 77.64). BERTurk leads by 1.46
points, and the source of that difference is concentrated in two tasks: CoLA
(41.71 vs 37.63) and STS-B (72.08 vs 69.59) both known weak spots of an
NSP-free architecture (Section~\ref{sec:cola}).

\paragraph{Notes on variance.} The reliable tasks are \texttt{qqp} ($\pm$0.07),
\texttt{mnli} ($\pm$0.19), \texttt{qnli} ($\pm$0.17) and \texttt{sst2}
($\pm$0.19); there, even 0.1-point differences are meaningful. TabiBERT's
$\pm$5.38 on STS-B is a genuine instability persisting across all five seeds
almost the entirety of that model's total deviation originates here.

\subsection{TabiBench}
\label{sec:tabibench}
TabiBench~\citep{tabibert} spans 28 datasets in eight categories and reaches
beyond sentence-level understanding into retrieval, code and academic text. It
is therefore complementary to TrGLUE rather than a substitute: it covers more
capabilities, but reports single-seed scores.

\begin{table}[H]
\centering
\footnotesize
\setlength{\tabcolsep}{3.5pt}
\caption{TabiBench category scores. Metrics by category: macro-F1 (text
classification, NLI, academic), micro-F1 at word level (token classification),
Pearson (STS), F1 (QA), NDCG@10 (both retrieval categories). Bold: best in
column among monolingual Turkish models. Reference scores for the four models
below the rule are taken from~\citet{tabibert}, Table~4; the ModernBERT-TR row
is read from the figures of its release article~\citep{modernberttr}, which
does not publish a numeric table. mmBERT is a 307M-parameter multilingual model
trained on 3T tokens and is listed as a reference point, not as a comparable
system.}
\label{tab:tabibench}
\begin{tabular}{lrrrrrrrrr}
\toprule
\textbf{Model} & \textbf{Text} & \textbf{Token} & \textbf{STS} & \textbf{NLI} &
\textbf{QA} & \textbf{Acad.} & \textbf{IR} & \textbf{Code} & \textbf{Total} \\
\midrule
\ours \textbf{MoganBert-TR} & 83.71 & 91.20 & 85.43 & 84.14 & 68.37 &
\textbf{72.62} & 75.78 & \textbf{60.57} & 77.73 \\
\midrule
ModernBERT-TR & \textbf{85.21} & 90.78 & \textbf{86.08} & \textbf{84.74} & 67.03 & 71.83 & \textbf{77.51} & 60.16 & \textbf{77.92} \\
TabiBERT & 83.44 & 93.42 & 84.74 & 84.51 & \textbf{69.71} & 72.44 & 75.44 & 56.95 & 77.58 \\
BERTurk & 83.42 & \textbf{93.67} & 85.33 & 84.33 & 60.16 & 71.40 & 74.84 & 54.54 & 75.96 \\
YTU-Cosmos-BERT~\citep{ytucosmos} & 84.25 & 93.60 & 84.68 & 84.16 & 31.50 & 71.78 & 74.29 & 53.80 & 72.26 \\
TurkishBERTweet~\citep{turkishbertweet} & 79.71 & 92.02 & 75.86 & 79.10 & 38.13 & 63.12 & 68.40 & 43.49 & 67.48 \\
\midrule
\multicolumn{10}{l}{\emph{Multilingual reference (not ranked)}} \\
mmBERT (307M) & 82.54 & 93.81 & 87.05 & 84.38 & 71.47 & 72.65 & 76.20 & 66.02 & 79.26 \\
\bottomrule
\end{tabular}
\end{table}

With 77.73, MoganBert-TR places second among monolingual Turkish encoders,
0.19 points behind ModernBERT-TR (77.92) and 0.15 points ahead of TabiBERT
(77.58). None of these three gaps should be read as an ordering. The benchmark's
authors state the constraint themselves: their scores come from single runs
without significance testing, margins below one point ``may or may not reach
statistical significance'', and they present the results as indicative rather
than as confirmed superiority~\citep{tabibert}. The MoganBert-TR scores are
likewise single-run, so the same caveat applies to them. The three models are
best described as tied at the top of the benchmark; what is informative is the
category structure underneath, where some differences are large enough to
survive that caveat.

\paragraph{Where MoganBert-TR leads.} The model ranks first among Turkish models
in two categories. The one that matters is \textbf{code retrieval} (60.57).
Against TabiBERT the margin is $+3.62$ and against BERTurk $+6.03$ far
outside any plausible single-run noise and it is the downstream counterpart
of two upstream decisions: the indentation-preserving tokenizer
(Section~\ref{sec:code-tok}) and the deliberately high 10\% code share of the
pretraining mixture. The per-task pattern is consistent with this: the gain over
TabiBERT concentrates on Apps-TR ($+4.90$), StackoverflowQA-TR ($+4.05$) and
CodeSearchNet-21K-TR ($+2.26$), all of which require matching Turkish natural
language against literal source code. Against ModernBERT-TR, however, the margin
is only $0.41$, and that model also allocates roughly a tenth of its mixture to
code; this suggests that the mixture share, rather than anything specific to
MoganBert-TR, is what separates both models from TabiBERT
(Section~\ref{sec:discussion}). The second lead, \textbf{academic understanding}
(72.62), is narrow ($+0.18$ over TabiBERT, $+0.79$ over ModernBERT-TR) and
driven by ThesisAbstract ($+2.67$), a 187-class problem over thesis abstracts.

\paragraph{Where MoganBert-TR trails.} \textbf{Token classification} is the
clearest weakness: 91.20 against 93.42 for TabiBERT and 93.67 for BERTurk, a
deficit of 2.2--2.5 points that is consistent across all four datasets
(WikiANN-TR $-2.63$, WikiNER $-1.34$, PosUD-IMST $-1.10$, PosUD-BOUN $-0.07$).
The uniformity of the sign across four datasets makes a measurement artefact
unlikely, but the deficit cannot be attributed: ModernBERT-TR scores 90.78 in
the same category with a different tokenizer, so a tokenizer-based explanation
does not follow from the available evidence, and the cause is left open.
\textbf{Question answering} (68.37) trails TabiBERT by 1.34 points, entirely on
TQuAD ($-2.35$); on XQuAD MoganBert-TR leads by 12.99 points, but with 179 test
examples that figure should not be over-read. In \textbf{STS}, \textbf{NLI} and
\textbf{information retrieval} ModernBERT-TR is ahead by 0.65, 0.60 and
1.73 points respectively.

\paragraph{Scope of the comparison.} The full 28-task protocol was run for
MoganBert-TR only. All reference scores are taken from the publications of the
models concerned rather than re-measured, because the hyperparameter search
alone 16 configurations per task across 28 tasks was beyond the compute
budget available to this project for six models. Consequently the MoganBert-TR
numbers and the reference numbers were produced in different environments, and
although the protocol is identical, environment-dependent variation cannot be
excluded. This is a further reason to read the small margins at the top of the
table as a tie.

Per-task scores are given in Appendix~\ref{app:tabibench}.

\subsection{CoLA: an architectural limit}
\label{sec:cola}
CoLA is the single largest item in the gap between MoganBert-TR and BERTurk.
All three ModernBERT models fall 4--15 points below BERTurk (BERTurk 41.71;
MoganBert-TR 37.63; TabiBERT 31.89; ModernBERT-TR 26.62), and MoganBert-TR is
the best among them.

The mechanism is architectural: in MLM, \texttt{[CLS]} is never masked and
therefore receives no direct gradient. In BERT the NSP loss is computed over
\texttt{[CLS]}, so that \texttt{[CLS]} is genuinely pretrained; ModernBERT
removed NSP which is why all three Turkish ModernBERT models declare
\texttt{classifier\_pooling: mean}.

Seven configurations were tried (lr 2e-5/3e-5/5e-5/1e-4/1e-3, batch 32/128,
cls/mean pooling), and the same pattern emerged along three independent axes:
\emph{the more intensive the training, the worse the result}. The lowest LR and
the shortest training give the best result; at lr 1e-3 the model collapses
entirely ($-$0.25). Opening up 149M parameters to 7,916 examples means
overfitting, and CoLA is the task most sensitive to it. The cause of the
collapse is a scale mismatch: pretraining used 8e-4 but with a global batch of
2,097,152 tokens, whereas CoLA fine-tuning uses a batch of 16,384 tokens
(128$\times$ smaller) with no warmup.

\subsection{Raw embedding geometry}
\label{sec:geometry}
The raw representations of the pretrained model are not suitable for direct use
with cosine similarity. Anisotropy measured with mean pooling over the same 400
sentences:

\begin{table}[H]
\centering
\footnotesize
\caption{Anisotropy comparison. \texttt{cos\_raw} is the mean cosine similarity
of random sentence pairs; proximity to 1 means all vectors are collected within
a narrow cone.}
\label{tab:anisotropy}
\begin{tabular}{lrrr}
\toprule
\textbf{Model} & \textbf{\texttt{cos\_raw}} & \textbf{\texttt{eff\_rank}} &
\textbf{\texttt{top1\_var}} \\
\midrule
\ours MoganBert-TR & 0.9841 & 117.9 & 0.0824 \\
TabiBERT & 0.9405 & 125.1 & 0.0682 \\
ModernBERT-TR & 0.9234 & 125.5 & 0.0924 \\
BERTurk & \textbf{0.7848} & \textbf{126.8} & \textbf{0.0632} \\
\bottomrule
\end{tabular}
\end{table}

Anisotropy is present in all MLM encoders (0.92--0.98), but the highest value is
MoganBert-TR's; BERTurk is markedly better, and this is plausibly a side effect
of NSP (an unverified hypothesis). The direct consequence is that zero-shot
retrieval performance remains low: on a ten-set IR evaluation the raw model
scores NDCG@10 = 0.2361. This measurement is the point of departure for the
embedding model work in Section~\ref{sec:embed}.

\subsection{RoPE theta ablation}
Annealing raises the global theta to 160,000. Since TrGLUE uses
\texttt{max\_seq\_len}=128, whether this harms short sequences was tested
directly; the weights are identical and the only variable is theta in the
configuration.

\begin{table}[H]
\centering
\footnotesize
\caption{Effect of changing RoPE theta at inference time.}
\label{tab:rope}
\begin{tabular}{llrrr}
\toprule
\textbf{Task} & \textbf{Metric} & \textbf{theta 160k} & \textbf{theta 10k} & \textbf{Difference} \\
\midrule
stsb & pearson & 0.6933 & 0.6893 & $-$0.0040 \\
stsb & spearman & 0.6710 & 0.6698 & $-$0.0012 \\
mrpc & accuracy & 0.7150 & 0.7060 & $-$0.0090 \\
mrpc & f1 & 0.6570 & 0.6397 & $-$0.0173 \\
\bottomrule
\end{tabular}
\end{table}

All four measurements are unfavourable to theta 10k. Since the model was trained
with 160k, pulling it back to 10k at inference creates a frequency-base
mismatch; the theoretical disadvantage of a large theta on short sequences
remains small next to that mismatch. \textbf{Decision:}
\texttt{global\_rope\_theta}=160000; the 8192-context capability is retained at
no cost.

\section{Embedding Model}
\label{sec:embed}

The anisotropy measured in Section~\ref{sec:geometry} (\texttt{cos\_raw} =
0.9841) showed that the pretrained model cannot be used for retrieval in its raw
form. This section presents the two-phase route followed in order to produce
MoganBert-Embed, an embedding model separate from the encoder.
MTEB(Turkish)~\citep{mteb} was used as the measurement ground throughout (26
tasks, \texttt{max\_len} 2048, fp16).

\subsection{Phase 1 --- Teacher distillation}
Qwen3-Embedding-8B~\citep{qwen3emb} was used as the teacher (output dimension
3072). The corpus was given to the teacher without a task instruction. The loss
has two terms:
\[
L = 1.0 \cdot L_{\text{distill}} \;+\; 0.05 \cdot L_{\text{GOR}}
\]
The distillation is cosine-based and the student is projected \emph{up} from
768 to 3072 the teacher is not truncated; the student rises into its space.
GOR (global orthogonal regularization~\citep{gor}), meanwhile, is computed at
the 768-dimensional backbone output where the anisotropy was measured. In
feature distillation it is critical that the student sees the \emph{same input
format} as the teacher; if the formats diverge, the learned mapping shifts.

\begin{table}[H]
\centering
\footnotesize
\caption{Effect of Phase 1 on geometry.}
\label{tab:phase1}
\begin{tabular}{lrr}
\toprule
\textbf{Metric} & \textbf{Start (MLM)} & \textbf{After Phase 1} \\
\midrule
\texttt{cos\_raw} (anisotropy) & 0.9841 & \textbf{0.0851} \\
\texttt{eff\_rank} & 117.9 & \textbf{157.2} \\
\texttt{top1\_var} & 0.0824 & \textbf{0.0355} \\
Zero-shot IR (10 sets, NDCG@10) & 0.2361 & \textbf{0.5927} \\
\bottomrule
\end{tabular}
\end{table}

The fall of \texttt{cos\_raw} from 0.98 to 0.085 means the representations have
left the narrow cone and spread out into the space; retrieval rose from 0.2361
to 0.5927 at the same step.

\subsection{Phase 2 --- Contrastive fine-tuning}
\label{sec:phase2}
What differs between the versions of Phase 2 is not the model architecture but
which signal types enter the mixture and how each type is processed.

\paragraph{v1 --- a single signal type.} Retrieval pairs only:
\[
L = 1.0 \cdot L_{\text{NCE}} \;+\; 2.0 \cdot L_{\text{distill}}
     \;+\; 0.1 \cdot L_{\text{GOR}}
\]
The batch is split: the task half produces $L_{\text{NCE}}$ (with hard
negatives, $\tau$=0.02) and the corpus half produces $L_{\text{distill}}$ (the
Phase 1 anchor). $L_{\text{GOR}}$ is computed \emph{separately} over the query
and document sets had it been computed jointly, it would push a query away
from its own positive document and work directly against $L_{\text{NCE}}$.
Result: IR 0.5927 $\rightarrow$ 0.6786.

\paragraph{v2 --- negative design and two structural fixes.} v2 takes the
mixture beyond retrieval pairs. The decisive criterion is not the source itself
but whether an \emph{explicit} negative can be defined from that source. A
\textbf{mined} negative is selected from the pool by similarity: it is
topically close, and therefore hard, but carries no label guarantee. An
\textbf{explicit} negative comes from the structure of the data itself and its
opposition is guaranteed: the contradiction hypothesis of a premise, a
low-scoring sentence pair, or a sentiment sentence expressing the same topic
with the opposite polarity. When an explicit negative is available it is placed
in slot zero \emph{in place of} the mined one.

Two structural fixes were added. \emph{False-negative masking:} since a query
can have dozens of relevant documents, the other positives of that query within
a batch were marked as negatives, and InfoNCE~\citep{infonce} was teaching the
model that ``this relevant document is irrelevant''; all positives of the same
query were grouped and masked in the logits. \emph{Anchor decay:} a fixed
distillation weight of 2.0 tied the model to the Phase 1 geometry (the
validation cosine fell from 0.8896 to only 0.883 over 8,000 steps); the weight
was decayed to 0.25 but not reduced to zero the anchor is the only signal
holding up quality on general Turkish text not represented in the training
mixture.

\paragraph{v3 each signal type received its own loss.} In v2 all types were
fed into the same InfoNCE. In v3 three types are processed in accordance with
their own nature. \textbf{(i)} CoSENT~\citep{cosent} for STS: by thresholding,
v2 discarded the middle band ($\sim$40\% of the data) and ignored the ranking
information, whereas the STS metric is Spearman.
\[
L = \log\Big(1 + \sum_{s_i > s_j} e^{(\cos_j - \cos_i)/\tau}\Big)
\]
\textbf{(ii)} In NLI, contradiction was positioned as a \emph{hard} negative and
neutral as a \emph{soft} negative, and (hypothesis, premise) was added alongside
(premise, hypothesis) pair classification is a symmetric task.
\textbf{(iii)} A same-class mask was added for the label data: because the batch
is task-homogeneous, one third of the in-batch negatives in a three-class set
come from the same class, and at every step the model was being taught that
``these two positive reviews are not similar to each other''.

In addition, a set of scientific abstracts whose labels carry rhetorical roles
was removed from the mixture: two sentences with the same role are \emph{not}
semantically similar, and counting them as positives pushes the space towards
clustering by discourse function. The tendency of temperature-based sampling to
repeat small sources 30--38 times was also capped (at most 12 times the number
of pairs per source).

\paragraph{v4 and v5.} Two further runs were produced from the same recipe
family; they propose no new method, and their role is to serve as soup
components. v4 is almost equal to v3 in overall score (66.88 vs 66.97) but has a
different profile: it leads on retrieval and trails on pair classification, STS
and classification. Because components that remain close in weight space while
their error patterns diverge correct one another in the average, both were taken
into the soup. v5 was not evaluated individually and was not used, as it lowered
the result in the soup (67.85 vs 68.30).

\begin{table}[H]
\centering
\footnotesize
\caption{Phase 2 version results, MTEB(Turkish). v5 was not evaluated
individually.}
\label{tab:phase2}
\begin{tabular}{lrrrrrrr}
\toprule
\textbf{Version} & \textbf{OVERALL} & \textbf{Retr.} & \textbf{Pair} & \textbf{STS} &
\textbf{Class.} & \textbf{Clust.} & \textbf{Bitext} \\
\midrule
phase1 & 63.53 & 56.13 & 56.76 & 73.95 & 70.61 & \textbf{64.35} & 96.55 \\
phase2\_v2 & 64.56 & 57.95 & 57.56 & 70.59 & 71.34 & 64.18 & \textbf{98.72} \\
phase2\_v4 & 66.88 & \textbf{58.22} & 70.89 & 80.02 & 73.17 & 62.42 & 95.68 \\
\ours \textbf{phase2\_v3} & \textbf{66.97} & 57.77 & \textbf{71.39} & \textbf{80.94} & \textbf{73.78} & 62.32 & 95.66 \\
\bottomrule
\end{tabular}
\end{table}

Two of v3's three targeted changes clearly worked: pair classification 57.56
$\rightarrow$ 71.39 (+13.83) and STS 70.59 $\rightarrow$ 80.94 (+10.35).
Classification, at 71.34 $\rightarrow$ 73.78 (+2.44), fell short of the target.
Two areas regressed: clustering 64.18 $\rightarrow$ 62.32 and bitext 98.72
$\rightarrow$ 95.66. The soup phase exists precisely to close these two gaps.

\subsection{Sentiment-contrast data}
The most targeted signal of Phase 2 is a set of sentiment-contrast triples
produced from scratch (40,737 triples). In each record the \texttt{positive}
field expresses the same sentiment class in different words; in the
\texttt{negative} field the sentiment class has been changed while topic,
subject, context and length are held constant and as few words as possible are
altered not an unrelated sentence but a \emph{hard} counterexample. Neutral
was defined as a genuine class: an opinion is present but measured, mixed or
undecided. If there is \emph{no} opinion at all the record is discarded
entirely; texts such as news reports, definitions, recipes, announcements,
legislation and technical descriptions are dropped. The data is LLM-generated
and has not been human-validated (Section~\ref{sec:limitations}).

\subsection{Model soup}
\label{sec:soup}
A weighted average of the components was taken~\citep{soup}. Because checkpoints
descending from the same starting point (Phase 1) remain close to one another in
weight space, averaging is meaningful.

\begin{table}[H]
\centering
\footnotesize
\caption{Selected soups. ``Phase 1 share'' = the ratio of the Phase 1 weight to
the total weight.}
\label{tab:soups}
\begin{tabular}{lllrr}
\toprule
\textbf{Soup} & \textbf{Components} & \textbf{Weights} & \textbf{Phase 1 share} & \textbf{OVERALL} \\
\midrule
\ours \textbf{f34-heavy} & phase1+v3+v4 & 0.4 / 1 / 1 & \textbf{0.167} & \textbf{68.30} \\
f34-06 & phase1+v3+v4 & 0.6 / 1 / 1 & 0.231 & 68.22 \\
f-v3 & phase1+v3 & 0.5 / 1 & 0.333 & 68.22 \\
f34-05 & phase1+v3+v4 & 0.5 / 1 / 1 & 0.200 & 68.19 \\
\midrule
345 (no Phase 1) & v3+v4+v5 & equal & 0.000 & 67.08 \\
fisher-f34 & phase1+v3+v4 & 0.4 / 1 / 1 & --- & 67.19 \\
\bottomrule
\end{tabular}
\end{table}

Three findings stand out. \textbf{(1) Phase 1 must be in the mixture.} The soup
consisting only of Phase 2 versions gives the weakest result at 67.08; adding
Phase 1 raises it to 68.30. Phase 1 was individually the lowest-scoring model
(63.53), yet it brings clustering and bitext to the mixture (clustering 62.19
$\rightarrow$ 66.46, bitext 96.48 $\rightarrow$ 97.24) a direct compensation
for the two regressions of Phase 2. \textbf{(2) The sweet spot is $\sim$0.17}
and the curve is flat between 0.10 and 0.25 (67.08 / 68.07 / \textbf{68.30} /
68.19 / 68.22 / 68.22); what matters is that Phase 1 is \emph{present}, not its
exact ratio. \textbf{(3) Not every component that adds diversity helps:} adding
v2 drops the result to 68.10 and adding v5 to 67.85. Fisher-weighted
merging~\citep{fisher} also fell behind the plain average in every pairing
(67.19 vs 68.30); since the Fisher information matrix is estimated from a sample
of 1,500 sentences, it is plausible that estimation noise at that size
undermines the regularization the plain average provides.

\subsection{MTEB(Turkish) results}
The final soup was measured on the 26 tasks of MTEB(Turkish) under the same
conditions as five external references and the teacher model: fp16, cosine
similarity, at most 2048 tokens. The pooling and prompt convention of each model
was read from its own configuration. Qwen3-Embedding-8B is not a competitor in
this work but the teacher of the Phase 1 distillation; its score should be read
as a \emph{reference ceiling}.

\begin{table}[H]
\centering
\footnotesize
\setlength{\tabcolsep}{3.5pt}
\caption{MTEB(Turkish) category averages. ``Ctx'' = the model's position limit.
Bold: best in column among student models. The last row is the teacher model and
is not included in the ranking.}
\label{tab:mteb-kategori}
\begin{tabular}{lrrrrrrrrr}
\toprule
\textbf{Model} & \textbf{Param} & \textbf{Ctx} & \textbf{OVERALL} & \textbf{Retr.}
& \textbf{Pair} & \textbf{STS} & \textbf{Class.} & \textbf{Clust.} & \textbf{Bitext} \\
\midrule
\ours \textbf{MoganBert-Embed}$^{\ast}$ & \textbf{149M} & 8192 & \textbf{68.30} & 59.58 & \textbf{69.32} & 79.49 & 75.36 & \textbf{66.46} & 97.24 \\
ModernBERT-TR & 149M & 8192 & 68.13 & 59.41 & 69.21 & 77.61 & \textbf{76.51} & 63.20 & 94.04 \\
mE5-large-instruct & 560M & 512$^{\dagger}$ & 67.47 & \textbf{61.05} & 63.03 & \textbf{81.23} & 73.70 & 61.87 & \textbf{98.99} \\
BGE-M3 & 568M & 8192 & 67.36 & 60.26 & 68.72 & 79.60 & 72.79 & 60.80 & \textbf{98.99} \\
Mursit-Large-TR & 404M & 2048 & 62.00 & 55.70 & 55.19 & 74.60 & 68.66 & 61.53 & 86.72 \\
trmteb-tr-embed & 111M & 512$^{\dagger}$ & 59.71 & 52.54 & 56.81 & 74.96 & 70.62 & 63.17 & 37.75$^{\ddagger}$ \\
\midrule
\multicolumn{10}{l}{\emph{Teacher model (reference ceiling, not included in the ranking)}} \\
Qwen3-Embedding-8B & 7.57B & 40960 & 68.66 & 64.46 & 62.76 & 80.04 & 72.95 & 63.01 & 98.18 \\
\bottomrule
\end{tabular}

\vspace{3pt}
\raggedright
\scriptsize
$^{\ast}$ Weighted soup: Phase~1 + v3 + v4, ratio $0.4/1/1$.
$^{\dagger}$ The position table of these two models is limited to 512; since
they could not be measured at 2048, they were evaluated at 512.
$^{\ddagger}$ \texttt{trmteb-tr-embed} is monolingual; because WMT16 requires
cross-lingual matching, this task falls outside its scope.
\end{table}

\normalsize
Four observations stand out.

\textbf{1. The student reaches 99.5\% of its teacher.} The 149M-parameter model
trails its 7.57-billion-parameter teacher by 0.36 points with a 51$\times$
smaller backbone. The teacher's advantage is concentrated in retrieval (64.46 vs
59.58); in return, the student surpasses its teacher on pair classification
(69.32 vs 62.76), classification (75.36 vs 72.95) and clustering (66.46 vs
63.01). Distillation evidently does not transfer all capabilities in equal
measure and leaves a gap on capacity-sensitive tasks such as retrieval.

\textbf{2. MoganBert-Embed ranks first on the overall average among student
models} (68.30 vs ModernBERT-TR 68.13). The wins are scattered: across 26 tasks
the soup ranks first on 9, mE5 on 8, ModernBERT-TR on 4, BGE-M3 on 3 and Mursit
on 2.

\textbf{3. The difference is clearest on clustering.} MoganBert-Embed ranks
first on both clustering tasks and leads the runner-up by 3.26 points on the
category average; this is a direct consequence of the Phase 1 contribution
discussed in Section~\ref{sec:soup}.

\textbf{4. Retrieval is the relative weak spot.} The model ranks first on only
three of the 11 retrieval tasks and trails mE5 and BGE-M3 on the category
average. The share of the retrieval signal in the Phase 2 mixture had been
reduced in favour of other signal types with v3; this result is the price of
that choice.

The full task-level table is given in Appendix~\ref{app:mteb}.

\section{Discussion}
\label{sec:discussion}

\paragraph{The training objective is a lever independent of the architecture.}
The Turkish encoder literature has so far focused on updating the architecture;
the measurements reported here show that, under the same architecture and the
same compute budget, the \emph{objective} alone can produce a qualitative
difference. That difference, moreover, is not uniform: it is invisible on
classification probes and grows threefold on cosine-based tasks. This is a
concrete warning about what is missed when encoder comparisons rest on
GLUE-style benchmarks alone.

\paragraph{The annealing phase is a cheap design space.} Branching produces two
different regimes for only $\sim$4.3\% of the total cost and, in these
measurements, gives a better result than a model soup. The role of context
length during decay turned out to be as decisive as that of the data mixture;
this extends mmBERT's~\citep{mmbert} decay-phase finding along a complementary
axis.

\paragraph{The code-retrieval gap tracks the mixture, not the objective.} Code
retrieval is the category in which MoganBert-TR separates most clearly from
TabiBERT ($+3.62$) and BERTurk ($+6.03$), and the natural question is which
design decision produced it. Two candidates are the indentation-preserving
tokenizer, which reaches full lossless roundtrip on the code corpus against
49\% for TabiBERT (Section~\ref{sec:code-tok}), and the code share of the
pretraining mixture, 10\% against TabiBERT's 6\%. A third possibility that
initially seems plausible is the CLM phase: source code is more strictly
sequential than natural language, so a left-to-right objective might capture
its structure better than masked prediction over a 30\%-corrupted sequence.

The comparison with ModernBERT-TR argues against that third explanation. That
model is trained with masked language modelling only, has no CLM phase, uses a
different tokenizer, and allocates roughly a tenth of its mixture to
code~\citep{modernberttr} and it scores 60.16, within half a point of
MoganBert-TR's 60.57, while TabiBERT with a 6\% code share sits nearly four
points below both. The variable that co-varies with the outcome is the mixture
share, not the objective and not the tokenizer family. No code-specific benefit
is therefore claimed for the curriculum; the retrieval gain measured for it
(Section~\ref{sec:ablation}) was observed on natural-language retrieval, and
nothing in these experiments extends it to code. Two models are a small sample
and they differ in many other respects, so this is an observation rather than a
controlled result; isolating the factors would require training the same
mixture under two tokenizers and the same tokenizer under two objectives.

\paragraph{Small-sample internal metrics are misleading.} Although the model
soup beat both of its parents on the fill-mask probe, it ranked fifth of six
models on TrGLUE. Basing checkpoint selection on small probes is common
practice; this case demonstrates its cost.

\paragraph{There is an architectural ceiling.} That BERTurk leads on CoLA and
STS-B is no accident: removing the NSP loss leaves the \texttt{[CLS]}
representation unpretrained. That all three Turkish ModernBERT models fall
behind in the same direction supports this reading. A lightweight auxiliary
objective providing a direct gradient to \texttt{[CLS]} during pretraining is an
unmeasured but promising direction.

\section{Limitations}
\label{sec:limitations}
\begin{itemize}[leftmargin=1.4em]
\item \textbf{The objective ablation is single-seeded} and rests on short runs
of 10,000 steps; whether the advantage is preserved at full-pretraining scale
must be verified separately. The retrieval leg rests on a single dataset, and
two of the probe tasks are saturated.
\item \textbf{The 16.6\% CLM ratio used in full pretraining has not been
validated by ablation;} the ablation result holds for 25\%. The ``pure 1024
annealing'' branch has likewise not been measured.
\item \textbf{The soup components are single-seeded,} and because the soup
weights were swept on MTEB they carry some degree of overfitting; the flatness
of the curve between 0.10 and 0.25 limits this risk but does not remove it.
\item \textbf{TabiBench is single-seed, and only MoganBert-TR was run in this
work.} Both the MoganBert-TR scores and the reference scores come from one run
per task, and the reference scores were produced in a different environment;
the benchmark's authors already caution that margins below one point may not be
significant~\citep{tabibert}. The 0.19-point gap to ModernBERT-TR and the
0.15-point gap to TabiBERT are therefore not an ordering. Re-measuring all six
models in one environment would settle this, but the hyperparameter search it
requires was outside the available compute budget.
\item \textbf{Two reference models were measured at 512 tokens} and the others
at 2048; on retrieval tasks involving long documents this works against those
two models, and their scores should be read as lower bounds.
\item \textbf{A single teacher was used,} and the sentiment data is
LLM-generated and has not been human-validated.
\item \textbf{Because of licensing restrictions} the corpus cannot be released
in full; the filtering decisions and the pipeline code are, however, shared for
reproducibility.
\end{itemize}

\section{Conclusion}
\label{sec:conclusion}
This work shows that an encoder foundation model trained from scratch for
Turkish from data collection through to an embedding model can be
produced by a small team without institutional infrastructure. The main finding
is that the choice of objective in encoder pretraining is decisive for Turkish
as well: under an equal step budget, the \clmmlm{} curriculum outperformed pure
MLM by 2.7--3.7$\times$ on cosine-based tasks, and its mechanism was measured in
embedding geometry. The second finding is that the annealing phase is a cheap
but effective design space: branching the decay after a shared prefix gave a
result 0.75 TrGLUE points better than a model soup at $\sim$4.3\% additional
cost.

MoganBert-TR is the best of the Turkish ModernBERT models compared, with
78.41 on TrGLUE, and reaches 77.73 on TabiBench, statistically indistinguishable
from ModernBERT-TR (77.92) and TabiBERT (77.58) at the top of that benchmark;
its clearest category-level advantage is code retrieval, where the mixture and
tokenizer decisions of Sections~\ref{sec:code-tok} and~\ref{sec:phaseplan}
translate into a 3.62-point margin over TabiBERT. MoganBert-Embed, the embedding
model derived from it, ranks first among student models on MTEB(Turkish)
with 68.30.

\paragraph{Future work.} There are three priority directions. The
\emph{retrieval gap} can be closed by increasing the share of the retrieval
signal in the Phase 2 mixture or by producing a separate retrieval-focused soup
component. Along the \emph{scale} axis, a large variant ($\sim$400M) and a
decoder built on the same data pipeline and tokenizer would complete the family.
As for \emph{ablation gaps}, the CLM ratio used in full pretraining and the pure
1024 annealing branch should be tested in a controlled manner, and the
code-retrieval advantage should be decomposed into its tokenizer, mixture and
objective components (Section~\ref{sec:discussion}).
\section*{Availability}
Model weights, the tokenizer, the embedding model and the evaluation code will
be released openly at \url{https://huggingface.co/moganai} upon publication of
this preprint. The data pipeline code will be shared together with the
filtering decisions so that the corpus can be reproduced; because of licensing
restrictions on part of the corpus, the corpus itself cannot be released in
full (Section~\ref{sec:limitations}).


\appendix
\newpage

\section{Data Pipeline Details}
\label{app:data}

\paragraph{Access and indexing.} Unauthenticated access to Turkish content in
Common Crawl is restricted at the index level; in practice the data can only be
downloaded via AWS S3 (\texttt{s3://commoncrawl}, us-east-1). For each monthly
crawl, Turkish content is first indexed from that month's columnar index
(Parquet) according to the \texttt{content\_languages} field; this step extracts
the triple \texttt{warc\_filename}, \texttt{warc\_record\_offset} and
\texttt{warc\_record\_length}. The download rate at production scale is
$\sim$450--480 records/second (64 concurrent workers, c7i.8xlarge).

\paragraph{Filter chain.} WARC parsing, boilerplate removal, Gopher/C4 quality
filters, Turkish boilerplate check, language verification, fastText quality
classification and PII masking (Turkish national ID checksum verification,
telephone, e-mail, IBAN, and the Luhn algorithm for credit cards).

\paragraph{Quality classifier details.} The LLM rubric was calibrated so that
gambling/casino/adult content would receive a fixed 0 and
advertising/real-estate-listing/SEO content would stay below 3. The threshold
choice was corrected empirically: when tested with
\texttt{keep\_labels}=\{3,4,5\}, the discarded \texttt{label}=2 documents turned
out in large majority to be genuine news, stock-market, recipe and article
content (words such as ``bet'' or ``fiyat'' were mostly false positives:
``İstanbulspor'', ``Betondan''), and \texttt{label}=2 was therefore added to the
accepted set. The distilled fastText was trained with \texttt{lr}=1.0,
\texttt{epoch}=25, \texttt{wordNgrams}=2, \texttt{dim}=100,
\texttt{loss}=softmax.

\paragraph{Resource and worker optimization.} The fastText quality model
($\sim$2.29 GB) is loaded separately in each parallel worker. On c7i.8xlarge
(32 vCPU, 61 GB RAM), OOM was observed at 30 and 24 workers; at 16 workers
memory use remained stable ($\sim$23--30 GB).

\paragraph{Ablation subset.} CommonCrawl 3.63B, Code 1.70B, FineWeb-edu-eng
1.25B, Legal 780M, Articles 627M, FinePdfs 588M, Books 316M, FineMath-4Plus
303M, turkce-matematik 221M $\sim$9.4B tokens in total.

\section{Tokenizer Implementation Details}
\label{app:tokenizer}

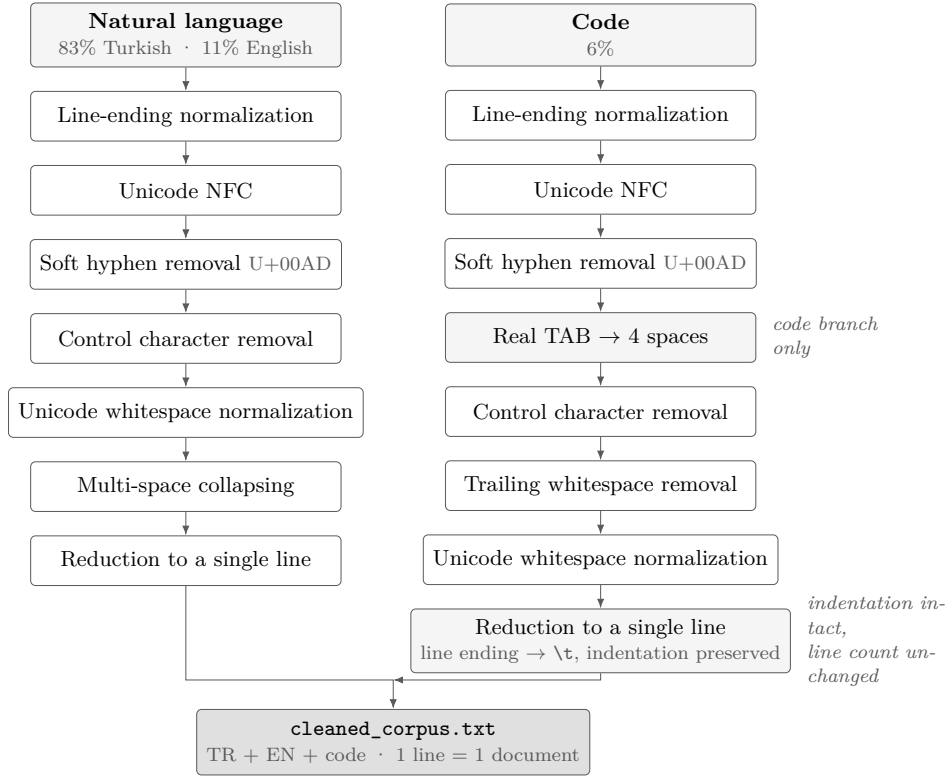
\begin{figure}[H]
\centering
\resizebox{0.80\linewidth}{!}{%
\begin{tikzpicture}[node distance=0.34cm]

\tikzset{
  pstep/.style = {draw=figline, line width=0.4pt, rounded corners=2pt,
                 fill=white, align=center, inner sep=4pt, font=\footnotesize,
                 minimum width=4.5cm, minimum height=0.72cm},
  phead/.style = {pstep, fill=figfill, minimum height=0.9cm, font=\footnotesize\bfseries},
}

\node[phead] (a0) {Natural language\\[-1pt]\textcolor{figline}{\scriptsize\mdseries 83\% Turkish · 11\% English}};
\node[pstep, below=of a0] (a1) {Line-ending normalization};
\node[pstep, below=of a1] (a2) {Unicode NFC};
\node[pstep, below=of a2] (a3) {Soft hyphen removal \textcolor{figline}{\scriptsize U+00AD}};
\node[pstep, below=of a3] (a4) {Control character removal};
\node[pstep, below=of a4] (a5) {Unicode whitespace normalization};
\node[pstep, below=of a5] (a6) {Multi-space collapsing};
\node[pstep, below=of a6] (a7) {Reduction to a single line};

\foreach \i/\j in {a0/a1,a1/a2,a2/a3,a3/a4,a4/a5,a5/a6,a6/a7}
  \draw[farrow] (\i) -- (\j);

\node[phead, right=1.5cm of a0] (b0) {Code\\[-1pt]\textcolor{figline}{\scriptsize\mdseries 6\%}};
\node[pstep, below=of b0] (b1) {Line-ending normalization};
\node[pstep, below=of b1] (b2) {Unicode NFC};
\node[pstep, below=of b2] (b3) {Soft hyphen removal \textcolor{figline}{\scriptsize U+00AD}};
\node[pstep, fill=figfill, below=of b3] (b4) {Real TAB $\rightarrow$ 4 spaces};
\node[pstep, below=of b4] (b5) {Control character removal};
\node[pstep, below=of b5] (b6) {Trailing whitespace removal};
\node[pstep, below=of b6] (b7) {Unicode whitespace normalization};
\node[pstep, fill=figfill, below=of b7] (b8)
  {Reduction to a single line\\[-1pt]\textcolor{figline}{\scriptsize line ending $\rightarrow$ \texttt{\textbackslash t}, indentation preserved}};

\foreach \i/\j in {b0/b1,b1/b2,b2/b3,b3/b4,b4/b5,b5/b6,b6/b7,b7/b8}
  \draw[farrow] (\i) -- (\j);

\coordinate (mid) at ($(a7.south)!0.5!(b8.south)$);
\node[pstep, fill=figaccent, below=1.15cm of mid, minimum width=5.4cm, minimum height=0.95cm] (merge)
  {\texttt{cleaned\_corpus.txt}\\[-1pt]\textcolor{figline}{\scriptsize TR + EN + code · 1 line = 1 document}};

\draw[farrow] (a7.south) |- ([yshift=0.42cm]merge.north) -- (merge.north);
\draw[farrow] (b8.south) |- ([yshift=0.42cm]merge.north);

\node[flab, right=0.12cm of b4, text width=2.3cm, align=left]
  {code branch\\only};
\node[flab, right=0.12cm of b8, text width=2.3cm, align=left]
  {indentation intact,\\line count unchanged};

\end{tikzpicture}}
\caption{The pre-tokenizer stage: separate but parallel normalization chains for
natural language and code data. Shaded boxes occur in the code branch only.}
\label{fig:pretok}
\end{figure}

\paragraph{The shared normalization pipeline.} In order: line-ending
normalization ($\backslash$r$\backslash$n, $\backslash$r $\rightarrow$
$\backslash$n), Unicode NFC, soft hyphen (U+00AD) removal, control character
removal (Cc/Cf; excluding $\backslash$n and $\backslash$t), Unicode whitespace
normalization and multi-space collapsing. The ordering is not arbitrary:
line-ending normalization must come before everything else. The rationale for
Unicode whitespace normalization rests on measurement in text scraped from
HTML sources, \texttt{\&nbsp;} (U+00A0) was very common (26 consecutive
occurrences in one legislative example; present in 99\% of records in the Court
of Cassation corpus). These characters are caught neither by the
\texttt{[ $\backslash$t]+} regex nor by Cc/Cf cleaning (they belong to the Zs
category), and if left uncleaned the tokenizer learns two versions of the same
word. Whitespace and punctuation splitting is not applied manually; it is left
to SentencePiece's own pre-tokenization.

\paragraph{The code branch.} There are two constraints: indentation must be
preserved and the line count must stay fixed. If indentation is removed, two
different Python programs collapse onto the same token sequence. If the line
endings of code are left as they are, the corpus balance also collapses: 521,000
code files amount to $\sim$15--52M lines, whereas all other sources together
amount to $\sim$900K lines, and SentencePiece samples on a per-line basis. The
solution is to escape the line ending to $\backslash$t; the obligatory
accompanying setting is \texttt{remove\_extra\_whitespaces}=False. This escape is
a technique for tokenizer training only in the pretraining data, code is
stored with real line endings.

\paragraph{Three silent problems encountered during training.} \textbf{(i)} The
default of \texttt{max\_sentence\_length} is 4192 bytes and every longer line is
silently discarded; since documents had been chunked with a target of $\sim$6 KB,
$\sim$75\% of them were being lost (solution: 100000). \textbf{(ii)} Because the
internal node index is a 32-bit signed int by default, training crashed once the
corpus reached 3.63 billion characters (solution:
\texttt{train\_extremely\_large\_corpus}=True). \textbf{(iii)} The normalizer
removes consecutive spaces, and
\texttt{normalization\_rule\_name="identity"} does not disable this (it is a
separate field).

\paragraph{Indentation tokens.} \texttt{allow\_whitespace\_only\_pieces} is
\texttt{False} by default and forbids learning pieces consisting of whitespace
only. When enabled, the token count for a five-line Python function drops from
46 to 30 (the lossy setting that removes indentation gives 26); in a real file,
5,612 characters of code fell from 2,535 to 2,290 tokens.

\paragraph{Special token layout.} \texttt{[UNK]}\,=\,0, \texttt{[PAD]}\,=\,1,
\texttt{[CLS]}\,=\,2, \texttt{[SEP]}\,=\,3, \texttt{[MASK]}\,=\,4,
\texttt{[EOS]}\,=\,5, \texttt{[unused0..15]}=6--21, byte tokens
(\texttt{byte\_\allowbreak{}fallback}=True) 22--277, learned pieces 278+. The
built-in BOS/EOS mechanism was disabled (\texttt{bos\_id}=$-1$,
\texttt{eos\_id}=$-1$); otherwise the vocabulary would contain two sets of
tokens for the same job.

\paragraph{Two rejected proposals.} Adding Turkish suffixes as
\texttt{user\_defined\_symbols}: this list does not respect word boundaries and,
if \emph{lar} were added, would forcibly cut through the middle of roots such as
``Dilara''; it is moreover unnecessary, as unigram learns frequent suffixes on
its own. The second is \texttt{normalization\_rule\_name="nmt\_nfkc"}: NFC is
already applied offline, and NFKC is more aggressive ($^2\rightarrow$2,
ﬁ$\rightarrow$fi), corrupting characters that carry meaning in a code/mathematics
corpus. \texttt{split\_digits} was likewise not used; its main benefit appears in
generative tasks.

\paragraph{HuggingFace conversion.} There is deliberately no pre-tokenizer
had a Metaspace or Whitespace pre-tokenizer been used, consecutive spaces would
be split one by one and the learned indentation tokens could never be selected
during encoding. Control pieces were given a score of $-1000$: since
\texttt{tokenizers.Unigram} does not recognize type information, the lattice can
produce identifier 4 from a literal \texttt{[MASK]} string in the text; a
negative score prevents this, while atomic matching at inference is unaffected
because it comes from the \texttt{AddedToken} mechanism.

\section{Implementation: Four Silent-Failure Pitfalls and Memory}
\label{app:impl}

Since HuggingFace's ModernBERT implementation is bidirectional only, the
attention path was rewritten. Four pitfalls, each capable of silently producing
an incorrect model:

\begin{enumerate}[leftmargin=1.6em]
\item \textbf{Causal local window.} ModernBERT's local window is symmetric
($w/2$ left + $w/2$ right). In the CLM phase a symmetric window leaks the
future, and this is not visible in the loss the model silently copies. The
window must be $(w-1, 0)$ in causal mode and revert to $(w/2, w/2)$ in
bidirectional mode.
\item \textbf{Boundary-aware packing.} Documents are packed into fixed blocks,
but attention must not cross the document boundary (\texttt{cu\_seqlens}).
\item \textbf{Position-id resetting.} In a packed sequence, RoPE positions must
start from zero for each document; otherwise a short document at the end of the
pack receives the positional encoding of a long history it has never seen. In
this project positions come from a single source (the collator's explicit
\texttt{position\_ids}); FlashAttention's own internal rotary is deliberately
disabled.
\item \textbf{Document-boundary masking in the loss.} \texttt{cu\_seqlens} cuts
attention but does not cut the loss: in the shifted-label CLM loss, the last
token of the $i$-th document in the pack tries to predict the first token of the
$(i{+}1)$-th document. The label of each document's final position must be
$-100$.
\end{enumerate}

\paragraph{Memory.} In the CLM phase, $\sim$99\% of positions are labelled. In a
batch of $\sim$98,000 tokens the logits tensor alone amounts to $98\text{K}
\times 50{,}048$ in bf16 $\approx$ 9.8 GB, plus 19.5 GB with \texttt{.float()}
and a further 19.5 GB with \texttt{log\_softmax} it does not fit even on a
96 GB card. Plain chunking is not enough, because autograd stores the logits of
every chunk for backpropagation. With \texttt{torch.utils.checkpoint} the logits
are recomputed during backpropagation and peak memory falls to that of a single
chunk ($\sim$1.6 GB); equivalence was verified by test (difference 0.00e+00).

\section{TabiBench Task-Level Results}
\label{app:tabibench}

\begin{table}[H]
\centering
\scriptsize
\setlength{\tabcolsep}{3pt}
\caption{TabiBench, all 28 tasks and the eight category averages. Scores for
TabiBERT, BERTurk, YTU-Cosmos-BERT, TurkishBERTweet and mmBERT are taken
from~\citet{tabibert}, Tables~6--13. Bold: best in row among monolingual
Turkish models; mmBERT is excluded from the ranking, following the source work.
MB-TR denotes ModernBERT-TR, which publishes TabiBench results only at category
level~\citep{modernberttr}; its per-task scores are marked ``?''. Category
averages are weighted by test-set size; the overall score is the unweighted
macro average of the eight categories. Metrics are as listed in
Table~\ref{tab:tabibench}.}
\label{tab:tabibench-task}
\begin{tabular}{lrrrrrrrr}
\toprule
\textbf{Task} & \textbf{Test size} & \textbf{Mogan} & \textbf{Tabi} &
\textbf{BERTurk} & \textbf{YTU} & \textbf{TRBw} & \textbf{mmBERT} & \textbf{MB-TR} \\
\midrule
\multicolumn{9}{l}{\emph{Text classification}} \\
\quad NewsCat & 250 & \textbf{97.22} & 95.20 & 95.60 & 97.21 & 91.98 & 94.80 & ? \\
\quad BilTweetNews & 150 & 53.32 & 50.11 & \textbf{57.87} & 53.39 & 53.16 & 49.06 & ? \\
\quad GenderHateSpeech & 2{,}000 & 66.57 & 69.01 & 68.25 & \textbf{71.02} & 68.58 & 66.45 & ? \\
\quad ProductReviews & 35{,}275 & 84.72 & 84.32 & 84.30 & \textbf{85.04} & 80.37 & 83.51 & ? \\
\quad \emph{Weighted avg} & & 83.71 & 83.44 & 83.42 & \textbf{84.25} & 79.71 & 82.54 & 85.21 \\
\midrule
\multicolumn{9}{l}{\emph{Token classification}} \\
\quad WikiNER & 1{,}000 & 75.07 & 76.41 & \textbf{79.44} & 78.96 & 72.60 & 75.97 & ? \\
\quad WikiANN-TR & 10{,}000 & 92.71 & 95.34 & 95.37 & \textbf{95.41} & 94.05 & 95.85 & ? \\
\quad PosUD-BOUN & 979 & 90.33 & \textbf{90.40} & 89.82 & 89.19 & 89.74 & 90.72 & ? \\
\quad PosUD-IMST & 1{,}100 & 92.97 & 94.07 & \textbf{94.60} & 94.30 & 93.21 & 94.26 & ? \\
\quad \emph{Weighted avg} & & 91.20 & 93.42 & \textbf{93.67} & 93.60 & 92.02 & 93.81 & 90.78 \\
\midrule
\multicolumn{9}{l}{\emph{STS}} \\
\quad SICK-TR & 4{,}927 & \textbf{85.97} & 85.00 & 85.95 & 85.27 & 78.35 & 87.81 & ? \\
\quad STSb-TR & 1{,}379 & 83.49 & \textbf{83.84} & 83.12 & 82.55 & 66.96 & 84.31 & ? \\
\quad \emph{Weighted avg} & & \textbf{85.43} & 84.74 & 85.33 & 84.68 & 75.86 & 87.05 & 86.08 \\
\midrule
\multicolumn{9}{l}{\emph{NLI}} \\
\quad SNLI-TR & 9{,}824 & 86.18 & 86.47 & \textbf{87.21} & 87.18 & 83.05 & 86.44 & ? \\
\quad MultiNLI-TR & 4{,}923 & 80.06 & \textbf{80.60} & 78.57 & 78.15 & 71.21 & 80.28 & ? \\
\quad \emph{Weighted avg} & & 84.14 & \textbf{84.51} & 84.33 & 84.16 & 79.10 & 84.38 & 84.74 \\
\midrule
\multicolumn{9}{l}{\emph{Question answering}} \\
\quad TQuAD & 2{,}520 & 69.99 & \textbf{72.34} & 63.30 & 32.01 & 38.04 & 71.55 & ? \\
\quad XQuAD & 179 & \textbf{45.60} & 32.61 & 15.96 & 24.25 & 39.40 & 70.40 & ? \\
\quad \emph{Weighted avg} & & 68.37 & \textbf{69.71} & 60.16 & 31.50 & 38.13 & 71.47 & 67.03 \\
\midrule
\multicolumn{9}{l}{\emph{Academic}} \\
\quad PubmedRCT-20K-TR & 1{,}500 & 74.03 & 75.32 & \textbf{75.61} & 75.32 & 70.07 & 74.37 & ? \\
\quad SciCite-TR & 1{,}859 & 82.63 & 83.29 & 81.60 & \textbf{84.09} & 79.59 & 83.27 & ? \\
\quad ThesisAbstract-11K & 1{,}683 & \textbf{53.44} & 50.77 & 49.20 & 47.56 & 31.18 & 52.47 & ? \\
\quad MedNLI-TR & 1{,}422 & 80.76 & \textbf{80.85} & 79.90 & 80.62 & 72.05 & 80.83 & ? \\
\quad \emph{Weighted avg} & & \textbf{72.62} & 72.44 & 71.40 & 71.78 & 63.12 & 72.65 & 71.83 \\
\midrule
\multicolumn{9}{l}{\emph{Information retrieval}} \\
\quad BiText & 3{,}000 & \textbf{99.51} & 99.42 & 96.77 & 97.25 & 95.98 & 99.49 & ? \\
\quad MsMarco-TR & 31{,}692 & 83.29 & \textbf{83.31} & 81.73 & 81.99 & 74.20 & 83.68 & ? \\
\quad Scifact-TR & 339 & 79.37 & 74.22 & 78.47 & \textbf{79.88} & 68.25 & 80.00 & ? \\
\quad Fiqa-TR & 1{,}706 & 48.65 & 50.67 & 49.46 & \textbf{51.10} & 36.23 & 52.29 & ? \\
\quad NFCorpus-TR & 12{,}334 & \textbf{35.07} & 31.19 & 32.47 & 28.36 & 27.79 & 33.44 & ? \\
\quad Quora-TR & 15{,}675 & 90.95 & 92.47 & 92.72 & \textbf{92.87} & 86.84 & 92.78 & ? \\
\quad \emph{Weighted avg} & & \textbf{75.78} & 75.44 & 74.84 & 74.29 & 68.40 & 76.20 & 77.51 \\
\midrule
\multicolumn{9}{l}{\emph{Code retrieval}} \\
\quad Apps-TR & 3{,}770 & \textbf{22.39} & 17.49 & 18.18 & 14.47 & 4.48 & 32.24 & ? \\
\quad CosQA-TR & 500 & \textbf{88.54} & 88.11 & 85.72 & 82.04 & 82.32 & 89.31 & ? \\
\quad StackoverflowQA-TR & 1{,}994 & \textbf{86.90} & 82.85 & 79.93 & 82.65 & 67.01 & 90.43 & ? \\
\quad CodeSearchNet-21K-TR & 3{,}000 & \textbf{86.38} & 84.12 & 78.15 & 79.35 & 70.41 & 88.36 & ? \\
\quad \emph{Weighted avg} & & \textbf{60.57} & 56.95 & 54.54 & 53.80 & 43.49 & 66.02 & 60.16 \\
\midrule
\ours \textbf{OVERALL (macro)} & & \textbf{77.73} & 77.58 & 75.96 & 72.26 & 67.48 & 79.26 & 77.92 \\
\bottomrule
\end{tabular}
\end{table}

\paragraph{Data-loading corrections.} Three defects in the reference loaders
were found and fixed before measurement; all three distort scores silently
rather than raising errors, and are documented here for reproducibility.
\textbf{(i) WikiNER label mapping.} The loader derives label identifiers from
the split being loaded. When labels arrive as \texttt{ClassLabel} integers and a
split does not contain every label, the mapping shifts and \texttt{O} tokens are
decoded as entities. The label set was made independent of the split.
\textbf{(ii) MultiNLI test split.} The loader expects an unsplit
\texttt{validation\_matched} (9{,}815 rows) and divides it itself; since the
source dataset already ships it divided, the test set was halved. The two parts
were merged. \textbf{(iii) Code retrieval splits.} The loader looks for a split
named \texttt{dev}; the source datasets have since been re-uploaded under the
name \texttt{validation}, which silently triggered a fallback branch that
discarded three quarters of the training set (\texttt{apps\_tr}: 308 rows
instead of 1{,}235). The split name was corrected.

All 28 runs completed with a zero exit status, category membership matches the
reference work exactly with no task counted twice, and score scales are
consistent (QA reported on 0--100, all others on 0--1 and multiplied by 100 in
the tables).

\newpage
\section{MTEB(Turkish) Task-Level Results}
\label{app:mteb}

\begin{table}[H]
\centering
\scriptsize
\setlength{\tabcolsep}{3.5pt}
\caption{MTEB(Turkish), all 26 tasks. Bold: best in row among student models.
The rightmost column is the teacher model.}
\label{tab:mteb-gorev}
\begin{tabular}{lrrrrr|r}
\toprule
\textbf{Task} & \textbf{Mogan} & \textbf{MBERT-TR} & \textbf{mE5} &
\textbf{BGE-M3} & \textbf{Mursit} & \textbf{Qwen3 (teach.)} \\
\midrule
\multicolumn{7}{l}{\emph{Retrieval}} \\
\quad ArguAnaTR & \textbf{51.99} & 49.93 & 49.07 & 50.43 & 45.70 & 61.84 \\
\quad CQADupstackGamingTR & 54.96 & 56.36 & \textbf{61.29} & 58.41 & 53.60 & 67.08 \\
\quad FiQA2018TR & \textbf{47.85} & 46.20 & 47.82 & 45.00 & 40.65 & 53.35 \\
\quad MSMarcoTRRetrieval & 58.09 & 57.87 & 58.35 & 57.93 & \textbf{58.95} & 58.34 \\
\quad NFCorpusTR & \textbf{9.96} & 9.45 & 9.62 & 9.41 & 9.50 & 12.00 \\
\quad QuoraRetrievalTR & 95.80 & 95.18 & \textbf{95.91} & 95.82 & 95.05 & 95.75 \\
\quad SCIDOCSTR & 3.30 & 3.37 & \textbf{4.18} & 3.69 & 2.70 & 5.78 \\
\quad SciFactTR & 81.23 & 77.06 & \textbf{83.99} & 78.81 & 72.79 & 90.29 \\
\quad SquadTRRetrieval & 71.70 & 75.92 & \textbf{77.88} & 76.69 & 62.10 & 79.10 \\
\quad TQuadRetrieval & 86.43 & 87.13 & 87.25 & \textbf{90.52} & 81.56 & 89.30 \\
\quad XQuADRetrieval & 94.03 & 95.04 & \textbf{96.22} & 96.11 & 90.11 & 96.20 \\
\midrule
\multicolumn{7}{l}{\emph{Pair classification}} \\
\quad MnliTr & \textbf{67.33} & 65.92 & 63.88 & 66.82 & 54.09 & 61.31 \\
\quad SnliTr & 64.37 & \textbf{67.29} & 52.47 & 60.56 & 51.51 & 57.43 \\
\quad XNLI & 76.26 & 74.43 & 72.75 & \textbf{78.78} & 59.98 & 69.52 \\
\midrule
\multicolumn{7}{l}{\emph{STS}} \\
\quad STSbTR & 79.49 & 77.61 & \textbf{81.23} & 79.60 & 74.60 & 80.04 \\
\midrule
\multicolumn{7}{l}{\emph{Classification}} \\
\quad THYSentiment & \textbf{69.05} & 68.15 & 66.13 & 67.36 & 57.68 & 66.15 \\
\quad TSTimelineNewsCat & 64.39 & \textbf{67.91} & 65.02 & 64.60 & 63.08 & 63.48 \\
\quad Turkish75News & 93.33 & 93.33 & 92.67 & 88.00 & \textbf{96.67} & 94.67 \\
\quad TurkishIrony & 56.25 & 57.75 & \textbf{58.83} & 53.25 & 50.92 & 55.75 \\
\quad TurkishMovieSentiment & 86.08 & \textbf{89.34} & 85.25 & 86.64 & 72.63 & 83.38 \\
\quad TurkishNewsCategory & 89.36 & 93.64 & \textbf{93.80} & 90.00 & 90.00 & 92.60 \\
\quad TurkishOffensiveLang & \textbf{71.72} & 71.51 & 60.68 & 61.97 & 62.06 & 63.36 \\
\quad TurkishProductSentiment & \textbf{72.70} & 70.46 & 67.22 & 70.46 & 56.23 & 64.21 \\
\midrule
\multicolumn{7}{l}{\emph{Clustering}} \\
\quad TurkishAbstractCorpus & \textbf{67.93} & 62.02 & 61.94 & 59.39 & 60.20 & 63.73 \\
\quad TurkishColumnWriting & \textbf{64.99} & 64.38 & 61.80 & 62.21 & 62.86 & 62.29 \\
\midrule
\multicolumn{7}{l}{\emph{Bitext}} \\
\quad WMT16BitextMining & 97.24 & 94.04 & \textbf{98.99} & \textbf{98.99} & 86.72 & 98.18 \\
\midrule
\ours \textbf{OVERALL (26 tasks)} & \textbf{68.30} & 68.13 & 67.47 & 67.36 & 62.00 & 68.66 \\
\bottomrule
\end{tabular}
\end{table}


\begin{thebibliography}{31}
\providecommand{\natexlab}[1]{#1}
\providecommand{\url}[1]{\texttt{#1}}
\expandafter\ifx\csname urlstyle\endcsname\relax
  \providecommand{\doi}[1]{doi: #1}\else
  \providecommand{\doi}{doi: \begingroup \urlstyle{rm}\Url}\fi

\bibitem[Alkurdi et~al.(2026)Alkurdi, Kesgin, Yuce, and Amasyali]{modernberttr}
Besher Alkurdi, Himmet~Toprak Kesgin, Muzaffer~Kaan Yuce, and Mehmet~Fatih
  Amasyali.
\newblock {ModernBERT-TR}: A modern encoder foundation model for {Turkish}.
\newblock \emph{Research Square preprint}, 2026.
\newblock \doi{10.21203/rs.3.rs-9500377/v1}.

\bibitem[Altinok(2025)]{trglue}
Duygu Altinok.
\newblock Introducing {TrGLUE} and {SentiTurca}: A comprehensive benchmark for
  {Turkish} general language understanding and sentiment analysis.
\newblock \emph{arXiv preprint arXiv:2512.22100}, 2025.
\newblock \doi{10.48550/arXiv.2512.22100}.

\bibitem[Bayram et~al.(2025)Bayram, Fincan, G{\"u}m{\"u}{\c{s}}, Karaka{\c{s}},
  Diri, and Y{\i}ld{\i}r{\i}m]{trtok}
M.~Ali Bayram, Ali~Arda Fincan, Ahmet~Semih G{\"u}m{\"u}{\c{s}}, Sercan
  Karaka{\c{s}}, Banu Diri, and Sava{\c{s}} Y{\i}ld{\i}r{\i}m.
\newblock Tokenization standards for linguistic integrity: {Turkish} as a
  benchmark.
\newblock \emph{arXiv preprint arXiv:2502.07057}, 2025.
\newblock \doi{10.48550/arXiv.2502.07057}.

\bibitem[Broder(1997)]{minhash}
Andrei~Z. Broder.
\newblock On the resemblance and containment of documents.
\newblock In \emph{Proceedings of the Compression and Complexity of Sequences},
  1997.

\bibitem[Conneau et~al.(2020)Conneau, Khandelwal, Goyal, Chaudhary, Wenzek,
  Guzm{\'a}n, Grave, Ott, Zettlemoyer, and Stoyanov]{xlmr}
Alexis Conneau, Kartikay Khandelwal, Naman Goyal, Vishrav Chaudhary, Guillaume
  Wenzek, Francisco Guzm{\'a}n, Edouard Grave, Myle Ott, Luke Zettlemoyer, and
  Veselin Stoyanov.
\newblock Unsupervised cross-lingual representation learning at scale.
\newblock In \emph{Proceedings of ACL}, 2020.

\bibitem[Devlin et~al.(2019)Devlin, Chang, Lee, and Toutanova]{bert}
Jacob Devlin, Ming-Wei Chang, Kenton Lee, and Kristina Toutanova.
\newblock {BERT}: Pre-training of deep bidirectional transformers for language
  understanding.
\newblock In \emph{Proceedings of NAACL-HLT}, 2019.

\bibitem[Gisserot-Boukhlef et~al.(2025)Gisserot-Boukhlef, Boizard, Faysse,
  Alves, Malherbe, Martins, Hudelot, and Colombo]{clmmlm}
Hippolyte Gisserot-Boukhlef, Nicolas Boizard, Manuel Faysse, Duarte~M. Alves,
  Emmanuel Malherbe, Andr{\'e} F.~T. Martins, C{\'e}line Hudelot, and Pierre
  Colombo.
\newblock Should we still pretrain encoders with masked language modeling?
\newblock \emph{arXiv preprint arXiv:2507.00994}, 2025.
\newblock \doi{10.48550/arXiv.2507.00994}.

\bibitem[H{\"a}gele et~al.(2024)H{\"a}gele, Bakouch, Kosson, Ben~Allal,
  Von~Werra, and Jaggi]{wsd}
Alexander H{\"a}gele, Elie Bakouch, Atli Kosson, Loubna Ben~Allal, Leandro
  Von~Werra, and Martin Jaggi.
\newblock Scaling laws and compute-optimal training beyond fixed training
  durations.
\newblock In \emph{Advances in Neural Information Processing Systems}, 2024.

\bibitem[Hinton et~al.(2015)Hinton, Vinyals, and Dean]{distill}
Geoffrey Hinton, Oriol Vinyals, and Jeff Dean.
\newblock Distilling the knowledge in a neural network.
\newblock \emph{arXiv preprint arXiv:1503.02531}, 2015.

\bibitem[Kesgin et~al.(2023)Kesgin, Yuce, and Amasyali]{ytucosmos}
Himmet~Toprak Kesgin, Muzaffer~Kaan Yuce, and Mehmet~Fatih Amasyali.
\newblock Developing and evaluating tiny to medium-sized {Turkish} {BERT}
  models.
\newblock \emph{arXiv preprint arXiv:2307.14134}, 2023.
\newblock \doi{10.48550/arXiv.2307.14134}.

\bibitem[Kudo and Richardson(2018)]{sentencepiece}
Taku Kudo and John Richardson.
\newblock {SentencePiece}: A simple and language independent subword tokenizer
  and detokenizer for neural text processing.
\newblock In \emph{Proceedings of EMNLP: System Demonstrations}, 2018.

\bibitem[Lozhkov et~al.(2024)Lozhkov, Ben~Allal, von Werra, and
  Wolf]{finewebedu}
Anton Lozhkov, Loubna Ben~Allal, Leandro von Werra, and Thomas Wolf.
\newblock {FineWeb-Edu}: The finest collection of educational content.
\newblock Hugging Face, 2024.

\bibitem[Marone et~al.(2025)Marone, Weller, Fleshman, Yang, Lawrie, and
  Van~Durme]{mmbert}
Marc Marone, Orion Weller, William Fleshman, Eugene Yang, Dawn Lawrie, and
  Benjamin Van~Durme.
\newblock {mmBERT}: A modern multilingual encoder with annealed language
  learning.
\newblock \emph{arXiv preprint arXiv:2509.06888}, 2025.

\bibitem[Matena and Raffel(2022)]{fisher}
Michael Matena and Colin Raffel.
\newblock Merging models with fisher-weighted averaging.
\newblock In \emph{Advances in Neural Information Processing Systems}, 2022.

\bibitem[Muennighoff et~al.(2023)Muennighoff, Tazi, Magne, and Reimers]{mteb}
Niklas Muennighoff, Nouamane Tazi, Lo{\"i}c Magne, and Nils Reimers.
\newblock {MTEB}: Massive text embedding benchmark.
\newblock In \emph{Proceedings of EACL}, 2023.

\bibitem[Najafi and Varol(2023)]{turkishbertweet}
Ali Najafi and Onur Varol.
\newblock {TurkishBERTweet}: Fast and reliable large language model for social
  media analysis.
\newblock \emph{arXiv preprint arXiv:2311.18063}, 2023.
\newblock \doi{10.48550/arXiv.2311.18063}.

\bibitem[{NLLB Team}(2022)]{flores}
{NLLB Team}.
\newblock No language left behind: Scaling human-centered machine translation.
\newblock \emph{arXiv preprint arXiv:2207.04672}, 2022.

\bibitem[Penedo et~al.(2024)Penedo, Kydl{\'\i}{\v{c}}ek, Ben~Allal, Lozhkov,
  Mitchell, Raffel, Von~Werra, and Wolf]{fineweb}
Guilherme Penedo, Hynek Kydl{\'\i}{\v{c}}ek, Loubna Ben~Allal, Anton Lozhkov,
  Margaret Mitchell, Colin Raffel, Leandro Von~Werra, and Thomas Wolf.
\newblock The {FineWeb} datasets: Decanting the web for the finest text data at
  scale.
\newblock In \emph{Advances in Neural Information Processing Systems (Datasets
  and Benchmarks Track)}, 2024.

\bibitem[Penedo et~al.(2025)Penedo, Kydl{\'\i}{\v{c}}ek, {\v{S}}abolcec,
  Messmer, Foroutan, Kargaran, Raffel, Jaggi, Von~Werra, and Wolf]{fineweb2}
Guilherme Penedo, Hynek Kydl{\'\i}{\v{c}}ek, Vinko {\v{S}}abolcec, Bettina
  Messmer, Negar Foroutan, Amir~Hossein Kargaran, Colin Raffel, Martin Jaggi,
  Leandro Von~Werra, and Thomas Wolf.
\newblock {FineWeb2}: One pipeline to scale them all adapting pre-training
  data processing to every language.
\newblock \emph{arXiv preprint arXiv:2506.20920}, 2025.

\bibitem[{Qwen Team}(2025)]{qwen3emb}
{Qwen Team}.
\newblock {Qwen3 Embedding}: Advancing text embedding and reranking through
  foundation models.
\newblock \emph{arXiv preprint arXiv:2506.05176}, 2025.

\bibitem[Schweter(2020)]{berturk}
Stefan Schweter.
\newblock {BERTurk} {BERT} models for {Turkish}.
\newblock Zenodo, doi:10.5281/zenodo.3770924, 2020.

\bibitem[Shah et~al.(2024)Shah, Bikshandi, Zhang, Thakkar, Ramani, and
  Dao]{flashattn}
Jay Shah, Ganesh Bikshandi, Ying Zhang, Vijay Thakkar, Pradeep Ramani, and Tri
  Dao.
\newblock {FlashAttention-3}: Fast and accurate attention with asynchrony and
  low-precision.
\newblock In \emph{Advances in Neural Information Processing Systems}, 2024.

\bibitem[Su(2022)]{cosent}
Jianlin Su.
\newblock {CoSENT}~(i): A more effective sentence embedding scheme than
  {Sentence-BERT}.
\newblock Scientific Spaces (blog), \url{https://kexue.fm/archives/8847}, 2022.

\bibitem[Su et~al.(2024)Su, Ahmed, Lu, Pan, Bo, and Liu]{rope}
Jianlin Su, Murtadha Ahmed, Yu~Lu, Shengfeng Pan, Wen Bo, and Yunfeng Liu.
\newblock {RoFormer}: Enhanced transformer with rotary position embedding.
\newblock \emph{Neurocomputing}, 568:\penalty0 127063, 2024.

\bibitem[Timkey and van Schijndel(2021)]{rogue}
William Timkey and Marten van Schijndel.
\newblock All bark and no bite: Rogue dimensions in transformer language models
  obscure representational quality.
\newblock In \emph{Proceedings of EMNLP}, 2021.

\bibitem[T{\"u}rker et~al.(2025)T{\"u}rker, K{\i}z{\i}lo{\u{g}}lu,
  G{\"u}ng{\"o}r, and {\"U}sk{\"u}darl{\i}]{tabibert}
Melik{\c{s}}ah T{\"u}rker, Asude~Ebrar K{\i}z{\i}lo{\u{g}}lu, Onur
  G{\"u}ng{\"o}r, and Susan {\"U}sk{\"u}darl{\i}.
\newblock {TabiBERT}: A large-scale {ModernBERT} foundation model and a unified
  benchmark for {Turkish}.
\newblock \emph{arXiv preprint arXiv:2512.23065}, 2025.

\bibitem[van~den Oord et~al.(2018)van~den Oord, Li, and Vinyals]{infonce}
Aaron van~den Oord, Yazhe Li, and Oriol Vinyals.
\newblock Representation learning with contrastive predictive coding.
\newblock \emph{arXiv preprint arXiv:1807.03748}, 2018.

\bibitem[Warner et~al.(2024)Warner, Chaffin, Clavi{\'e}, Weller, Hallstr{\"o}m,
  Taghadouini, Gallagher, Biswas, Ladhak, Aarsen, Cooper, Adams, Howard, and
  Poli]{modernbert}
Benjamin Warner, Antoine Chaffin, Beno{\^i}t Clavi{\'e}, Orion Weller, Oskar
  Hallstr{\"o}m, Said Taghadouini, Alexis Gallagher, Raja Biswas, Faisal
  Ladhak, Tom Aarsen, Nathan Cooper, Griffin Adams, Jeremy Howard, and Iacopo
  Poli.
\newblock Smarter, better, faster, longer: A modern bidirectional encoder for
  fast, memory efficient, and long context finetuning and inference.
\newblock \emph{arXiv preprint arXiv:2412.13663}, 2024.

\bibitem[Wenzek et~al.(2020)Wenzek, Lachaux, Conneau, Chaudhary, Guzm{\'a}n,
  Joulin, and Grave]{ccnet}
Guillaume Wenzek, Marie-Anne Lachaux, Alexis Conneau, Vishrav Chaudhary,
  Francisco Guzm{\'a}n, Armand Joulin, and Edouard Grave.
\newblock {CCNet}: Extracting high quality monolingual datasets from web crawl
  data.
\newblock In \emph{Proceedings of the Twelfth Language Resources and Evaluation
  Conference (LREC)}, 2020.

\bibitem[Wortsman et~al.(2022)Wortsman, Ilharco, Gadre, Roelofs, Gontijo-Lopes,
  Morcos, Namkoong, Farhadi, Carmon, Kornblith, and Schmidt]{soup}
Mitchell Wortsman, Gabriel Ilharco, Samir~Ya Gadre, Rebecca Roelofs, Raphael
  Gontijo-Lopes, Ari~S. Morcos, Hongseok Namkoong, Ali Farhadi, Yair Carmon,
  Simon Kornblith, and Ludwig Schmidt.
\newblock Model soups: Averaging weights of multiple fine-tuned models improves
  accuracy without increasing inference time.
\newblock In \emph{Proceedings of ICML}, 2022.

\bibitem[Zhang et~al.(2017)Zhang, Yu, Kumar, and Chang]{gor}
Xu~Zhang, Felix~X. Yu, Sanjiv Kumar, and Shih-Fu Chang.
\newblock Learning spread-out local feature descriptors.
\newblock In \emph{Proceedings of ICCV}, 2017.

\end{thebibliography}
\end{document}